\documentclass[runningheads]{llncs}

\usepackage{graphicx}
\usepackage{comment}
\usepackage{amsmath,amssymb}
\usepackage{color}
\usepackage{url}
\usepackage{hyperref}

\newif\ifreview
\reviewfalse

\ifreview
	\usepackage{lineno}

	\linenumbers
\fi

\usepackage{silence}
\usepackage{multirow}
\usepackage{colortbl}
\usepackage{color}
\usepackage{hyperref}
\usepackage{xspace}
\usepackage{tabularx}

\usepackage{tikz}
\usetikzlibrary{calc,shapes,arrows,backgrounds,positioning,backgrounds,fit}
\usepackage{pgfplots}
\pgfplotsset{compat=1.16}
\usepackage{pgfplotstable}
\usepackage{tikzscale}
\usepackage{csvsimple}
\usepackage{siunitx}
\usepackage{booktabs}
\usepackage{makecell}
\usepackage{eqnarray}
\usepackage{bbm}
\usepackage{textcomp}
\usepackage{gensymb}
\usepackage{pifont}
\usepackage{subcaption}
\usepackage{dsfont}
\usepackage{balance}
\usepackage{hyperref}
\usepackage[percent]{overpic}
\usepackage{bbding}
\usepackage{pict2e}
\usepackage{wrapfig}

\usepackage{algorithm}
\usepackage{algpseudocode}

\newcolumntype{Y}{>{\centering\arraybackslash}X}
\newcolumntype{Z}{>{\raggedleft\arraybackslash}X}

\newcommand{\refsec}[1]{Sec.~\ref{#1}}

\newcommand{\refeq}[1]{Eqn.~\ref{#1}}
\newcommand{\reffig}[1]{Fig.~\ref{#1}}
\newcommand{\reftab}[1]{Table~\ref{#1}}

\newcommand{\PAR}[1]{\vskip4pt \noindent {\bf #1~}}  

\newcommand*\widebar[1]{%
   \hbox{%
     \vbox{%
       \hrule height 0.5pt 
       \kern0.3ex
       \hbox{%
         \kern-0.1em
         \ensuremath{#1}%
         \kern-0.1em
       }%
     }%
   }%
} 

\def\eg{\emph{e.g.}\@\xspace}

\def\etal{\emph{et al.}\@\xspace}

\definecolor{scannetwall}              {RGB}{174, 199, 232}
\definecolor{scannetfloor}             {RGB}{152, 223, 138}
\definecolor{scannetcabinet}           {RGB}{ 31, 119, 180}
\definecolor{scannetbed}               {RGB}{255, 187, 120}
\definecolor{scannetchair}             {RGB}{188, 189,  34}
\definecolor{scannetsofa}              {RGB}{140,  86,  75}
\definecolor{scannettable}             {RGB}{255, 152, 150}
\definecolor{scannetdoor}              {RGB}{214,  39,  40}
\definecolor{scannetwindow}            {RGB}{197, 176, 213}
\definecolor{scannetbookshelf}         {RGB}{148, 103, 189}
\definecolor{scannetpicture}           {RGB}{196, 156, 148}
\definecolor{scannetcounter}           {RGB}{ 23, 190, 207}
\definecolor{scannetdesk}              {RGB}{247, 182, 210}
\definecolor{scannetcurtain}           {RGB}{219, 219, 141}
\definecolor{scannetrefridgerator}     {RGB}{255, 127,  14}
\definecolor{scannetshowercurtain}     {RGB}{158, 218, 229}
\definecolor{scannettoilet}            {RGB}{ 44, 160,  44}
\definecolor{scannetsink}              {RGB}{112, 128, 144}
\definecolor{scannetbathtub}           {RGB}{227, 119, 194}
\definecolor{scannetotherfurniture}    {RGB}{ 82,  84, 163}

\newcommand{\scannetcoltable}{
  \resizebox{\linewidth}{!}{%
    \begin{tikzpicture}[tight background, scale=0.75, every node/.style={font=\large}]
      \draw[black, fill=scannetwall, draw=white] (0,0) rectangle (1 * 4, 1) node[pos=0.5] {Wall};
      \draw[black, fill=scannetfloor, draw=white] (1 * 4,0) rectangle (2 * 4, 1) node[pos=0.5] {Floor};
      \draw[white, fill=scannetcabinet, draw=white] (2 * 4,0) rectangle (3 * 4, 1) node[pos=0.5] {Cabinet};
      \draw[black, fill=scannetbed, draw=white] (3 * 4,0) rectangle (4 * 4, 1) node[pos=0.5] {Bed};
      \draw[black, fill=scannetchair, draw=white] (4 * 4,-0) rectangle (5 * 4, 1) node[pos=0.5] {Chair};
      \draw[white, fill=scannetsofa, draw=white] (5 * 4,-0) rectangle (6 * 4, 1) node[pos=0.5] {Sofa};
      \draw[black, fill=scannettable, draw=white] (6 * 4,-0) rectangle (7 * 4, 1) node[pos=0.5] {Table};
      \draw[white, fill=scannetdoor, draw=white] (7 * 4,-0) rectangle (8 * 4, 1) node[pos=0.5] {Door};
      \draw[black, fill=scannetwindow, draw=white] (8 * 4,-0) rectangle (9 * 4, 1) node[pos=0.5] {Window};
      \draw[white, fill=scannetbookshelf, draw=white] (9 * 4,-0) rectangle (10 * 4, 1) node[pos=0.5] {Bookshelf};

      \draw[black, fill=scannetpicture, draw=white] (0 * 4,-1) rectangle (1 * 4, 0) node[pos=0.5] {Picture};
      \draw[white, fill=scannetcounter, draw=white] (1 * 4,-1) rectangle (4 * 2, 0) node[pos=0.5] {Counter};
      \draw[black, fill=scannetdesk, draw=white] (2 * 4,-1) rectangle (3 * 4, 0) node[pos=0.5] {Desk};
      \draw[black, fill=scannetcurtain, draw=white] (3 * 4,-1) rectangle (4 * 4, 0) node[pos=0.5] {Curtain};
      \draw[black, fill=scannetrefridgerator, draw=white] (4 * 4,-1) rectangle (5 * 4, 0) node[pos=0.5] {Refridgerator};
      \draw[black, fill=scannetshowercurtain, draw=white] (5 * 4,-1) rectangle (6 * 4, 0) node[pos=0.5] {Shower curtain};
      \draw[white, fill=scannettoilet, draw=white] (6 * 4,-1) rectangle (7 * 4, 0) node[pos=0.5] {Toilet};
      \draw[white, fill=scannetsink, draw=white] (7 * 4,-1) rectangle (8 * 4, 0) node[pos=0.5] {Sink};
      \draw[black, fill=scannetbathtub, draw=white] (8 * 4,-1) rectangle (9 * 4, 0) node[pos=0.5] {Bathtub};
      \draw[white, fill=scannetotherfurniture, draw=white] (9 * 4,-1) rectangle (10 * 4, 0) node[pos=0.5] {Other furniture};
    \end{tikzpicture}%
  }}
\definecolor{new_purple}{HTML}{D175F0}
\definecolor{new_gray}{HTML}{777777}
\definecolor{new_pink}{HTML}{FF69B4}
\definecolor{new_brown}{HTML}{8B4513}
\definecolor{new_blue_dark}{HTML}{1269B0}
\definecolor{new_red_dark}{HTML}{BF534F}
\definecolor{new_green_dark}{HTML}{1e942e}
\definecolor{new_yellow_dark}{HTML}{c99436}
\definecolor{new_cyan_dark}{HTML}{3ab7c7}
\definecolor{new_orange_dark}{HTML}{f05d02}

\definecolor{new_red}{HTML}{ea4335}
\definecolor{new_green}{HTML}{34a853}
\definecolor{new_blue}{HTML}{4285f4}
\definecolor{new_yellow}{HTML}{fbbc04}
\definecolor{new_orange}{HTML}{ff6d01}
\definecolor{new_lightblue}{HTML}{46bdc4}

\definecolor{gs_red}{HTML}{cc0000}
\definecolor{gs_blue}{HTML}{3d85c6}
\definecolor{gs_yellow}{HTML}{f1c232}
\definecolor{gs_green}{HTML}{6aa84f}
\definecolor{gs_orange}{HTML}{e69138}
\definecolor{gs_blue_light}{HTML}{3d85c6}

\begin{document}


\def\SubNumber{88}

\def\GCPRTrack{Fast Review Track}

\title{Global Hierarchical Attention for 3D Point Cloud Analysis}

\ifreview
\else

	\author{Dan Jia\orcidID{0000-0002-1174-7164} \and
	Alexander Hermans\orcidID{0000-0003-2127-0782} \and
	Bastian Leibe\orcidID{0000-0003-4225-0051}}
	
	\authorrunning{D. Jia et al.}
	
	\institute{Visual Computing Institute\\RWTH Aachen University\\
	\email{\{jia,hermans,leibe\}@vision.rwth-aachen.de}}
\fi

\maketitle              

\begin{abstract}
We propose a new attention mechanism, called Global Hierarchical Attention (GHA), for 3D point cloud analysis.
GHA approximates the regular global dot-product attention via a series of coarsening and interpolation operations over multiple hierarchy levels.
The advantage of GHA is two-fold.
First, it has linear complexity with respect to the number of points, enabling the processing of large point clouds.
Second, GHA inherently possesses the inductive bias to focus on spatially close points, while retaining the global connectivity among all points.
Combined with a feedforward network, GHA can be inserted into many existing network architectures.
We experiment with multiple baseline networks and show that adding GHA consistently improves performance across different tasks and datasets.
For the task of semantic segmentation, GHA gives a +1.7\%~mIoU increase to the MinkowskiEngine baseline on ScanNet.
For the 3D object detection task, GHA improves the CenterPoint baseline by +0.5\%~mAP on the nuScenes dataset, and the 3DETR baseline by +2.1\%~mAP$_{25}$ and +1.5\%~mAP$_{50}$ on ScanNet.
\end{abstract}

\section{Introduction}
\label{sec:introduction}

Recent years have witnessed a significant increase in interest for point cloud analysis~\cite{Qi17CVPR,Choy19CVPR,Thomas19ICCV}, which is fundamental in many applications.
Due to the lack of an inherent order in the point cloud representation, methods that aim to directly process a point cloud need to have permutation invariance~\cite{Qi17CVPR}.
Attention mechanisms, being a set to set operation (which is permutation invariant), are an interesting candidate for this task.
Originally introduced for language tasks, attention-based transformers have recently shown great success in a range of computer vision tasks~\cite{Dosovitskiy20ICLR,Carion20ECCV,Liu21ICCV,Caron21ICCV}.
Several recent works have applied transformer models to 3D tasks, including classification~\cite{Guo21CVM}, detection~\cite{Pan21CVPR,Liu21ICCVb,Misra21ICCV,Mao21ICCV}, and segmentation~\cite{Zhao21ICCV,Park21arXiv}, and have obtained promising results.

There are two major challenges when applying attention to point clouds.
Firstly, the required memory quickly grows out of bounds for large point clouds, due to the quadratic space complexity of the attention matrix.
Secondly, the global receptive field of an attention layer allows the network to easily overfit to distracting long range information, instead of focusing on local neighborhoods, which provide important geometric information~\cite{Raghu21arXiv,Choy19ICCV,Zeng17CVPR}.
To solve these two problems, existing point cloud methods often resort to computing attention only within a local neighborhood (\textit{local attention}), and channel global information either through customized multi-scale architectures~\cite{Zhao21ICCV,Pan21CVPR} or via modeling interactions between point clusters~\cite{Pan21CVPR}.

In this work we propose a novel attention mechanism, called \textit{Global Hierarchical Attention} (GHA), which natively addresses the aforementioned two challenges.
The memory consumption of GHA scales linearly with respect to the number of points, a significant reduction to the regular quadratic attention.
In addition, GHA by design embodies the inductive bias that encourages the network to focus more on local neighbors, while retaining a global receptive field.

The core of GHA is a series of hierarchical levels, connected using \textit{coarsening} and \textit{interpolation} operations.
Given an input point cloud, the coarsening operation is used to construct hierarchies with different levels of details.
Within each hierarchy level, local attention is computed, and the results from all levels are collated via the interpolation operation.
The output of GHA can be computed without needing to reconstruct the memory-intensive attention matrix, and it approximates the output of regular attention, while giving more emphasis to the nearby points.
Our method is inspired by the recent H-Transformer-1D~\cite{Zhu21ACL}, which produces an efficient hierarchical approximation of the attention matrix for 1D (language) sequences, but due to its reliance on the existence of a 1D order, it cannot be directly applied to point clouds.

We propose two realizations of GHA, compatible with either raw or voxelized point clouds.
We design an add-on module based on GHA that can be easily inserted into an existing network.
We experiment with various baseline networks on both 3D object detection and segmentation tasks.
Our experiments show that GHA brings a consistent improvement to a range of point cloud networks and tasks, including +1.7\%~mIoU on ScanNet segmentation for the MinkowskiEngine~\cite{Choy19CVPR}, +0.5\%~mAP on nuScenes detection for CenterPoint~\cite{Yin21CVPR}, and +2.1\%~mAP$_{25}$ and +1.5\%~mAP$_{50}$ on ScanNet detection for 3DETR.

\section{Related Work}
\label{sec:related_work}

\textbf{Point Cloud Analysis.}
Existing architectures for analyzing point clouds fall into several categories.
Projection-based methods~\cite{Kanezaki21PAMI,Lang19CVPR,Su18CVPR,Tatarchenko18CVPR} process the point clouds by projecting them onto 2D grids and applying 2D CNNs to the obtained image.
Voxel-based approaches~\cite{Graham18CVPR,Li18NIPS,Zhou17CVPR,Riegler17CVPR,Choy19CVPR} use 3D (sparse) CNNs on the voxelized point cloud.
In addition, there are methods that do not rely on discretization and operate directly on point clouds, via PointNet~\cite{Qi17NIPS}, continuous convolutions~\cite{Xu18ECCV,Thomas19ICCV,Wu19CVPR,Mao19ICCV},
or graph neural networks~\cite{Li21PAMI,Wang19CVPR,Wang18TOG,Zhao19CVPR,Landrieu18CVPR,Landrieu19CVPR,Jiang19ICCV}.
Recent works~\cite{Liu21ICCVb,Pan21CVPR,Guo21CVM,Zhao21ICCV,Fan21CVPR,Park21arXiv,Misra21ICCV} have attempted to adapt transformers and attention mechanisms~\cite{Vaswani17NIPS} for point cloud processing, encouraged by their recent success in language and vision tasks.

\noindent \textbf{Attention for Point Clouds.}
Several works have used attention mechanisms for point clouds.
Group-free Transformer~\cite{Liu21ICCVb} is an object detection framework that uses a PointNet++~\cite{Qi17NIPS} backbone to generate initial bounding box proposals, and a stack of self and cross-attention layers for proposal refinement.
Pointformer~\cite{Pan21CVPR} is another detection network which builds a multi-scale backbone network via a series of customized local and global attention mechanisms.
VoTr~\cite{Mao21ICCV} proposed to replace the sparse voxel convolution in the SECOND detector~\cite{Yan18Sensors} with local and dilated attention and obtained good detection results.  
These works rely on specially crafted attention mechanisms and network architectures.
3DETR~\cite{Misra21ICCV} is the first work that uses a standard end-to-end transformer architecture for object detection, following the framework laid out by the image-based DETR object detector~\cite{Carion20ECCV}.
In addition to detection, attention mechanisms have also been used on tasks like analyzing point cloud videos~\cite{Fan21CVPR}, registration~\cite{Wang19ICCVb}, or normal estimation~\cite{Guo21CVM}.

Two works have aimed at designing attention-based networks specifically for different point cloud tasks.
Zhao~\etal~\cite{Zhao21ICCV} proposed Point Transformer, a multi-scale U-net architecture with local attention computed among $k$-nearest-neighbors.
It was shown to perform well for shape classification on ModelNet40~\cite{Zhirong15CVPR}, object part segmentation on ShapeNet~\cite{Yi16TOG}, and semantic segmentation on S3DIS~\cite{Armeni16CVPR}.
Fast Point Transformer \cite{Park21arXiv} modified the Point Transformer architecture to operate on voxels for improved speed and performance.

Similar to these (\cite{Zhao21ICCV,Park21arXiv}), GHA is intended as a general-purpose method that can be applied to a wide range of tasks.
However, our focus lies on the attention mechanism itself, rather than the network architecture.
We conduct experiments by augmenting existing well-known networks with several attention layers, and compare the performance between networks using different types of attention (\eg GHA, local, or regular).
Excluding architecture design enables us to draw conclusions on the attention mechanism itself from comparative experiments.

\noindent \textbf{Efficient Transformers.} 
A large body of works exists in the NLP community that aims to reduce the memory consumption of the attention mechanism (\eg~\cite{Zaheer20NeurIPS,Choromanski21ICLR,Zhu21ACL,Wang20arXiv,Tay20arXiv}).
The most relevant to our work is that of Zhu and Soricut~\cite{Zhu21ACL}, which proposed a hierarchical attention mechanism to approximate the full attention matrix with linear space complexity.
As they exploit the fixed order of a 1D sequence to represent the attention matrix in a matrix block hierarchy, their exact formulation does not readily generalize to higher dimensions.
In this work, we extend this idea to point cloud analysis and propose the GHA mechanism that can work with both 3D point clouds and voxels.

In the vision community, most methods resort to computing attention only within local windows~\cite{Liu21ICCV,Zhao21ICCV} to reduce the memory consumption and rely on architectural level design to channel long-range information.
Compared to these methods, GHA has the advantage of having a global field of view, without imposing additional constraints on the network architecture.
As such, GHA can easily be integrated into a whole range of existing networks.

\section{Global Hierarchical Attention (GHA)}

\subsection{Background: Dot-product Attention}

Given the query, key, and value matrices $Q$, $K$, $V \in \mathbb{R}^{N \times d}$, the output $Z \in \mathbb{R}^{N \times d}$ of a regular scaled dot-product attention~\cite{Vaswani17NIPS} is defined as:
$Z = \text{softmax}( \frac{QK^{T}}{\sqrt{d}} ) V$,
where $Q$, $K$, $V$ represents the token queries, keys, values stacked as rows of the matrices, $N$ is the number of tokens, and $d$ is the dimension of the embedding space.
This equation can be expressed in a more compact matrix form:\\
$Z = D^{-1} A V$
with
$A = e^{S}$, 
$S_{ij} = \frac{Q_i K_j^{T}}{\sqrt{d}}$,
$D = \text{diag}( A \cdot \textbf{1}_N )$,
$\textbf{1}_N = [1, 1, \dots, 1]^T$ .\\
$S_{i, j}$ represents the unnormalized cosine similarity between the $i$-th query and the $j$-th key embedding.
The normalized attention matrix $D^{-1}A$ has $O(n^2)$ space complexity, making it memory intensive for large numbers of tokens.

\subsection{GHA for Point Cloud Analysis}
\label{sec:gha}

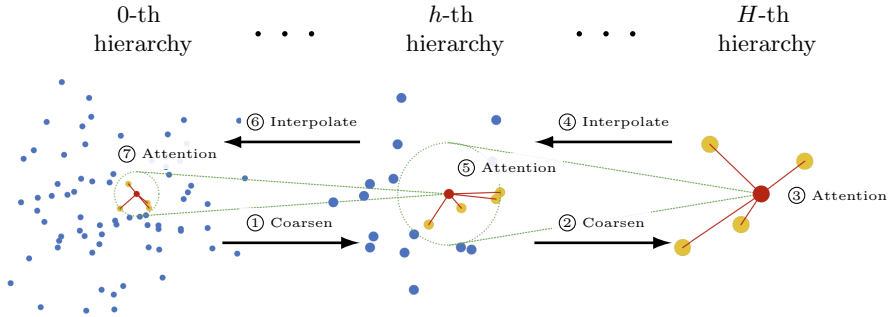
\begin{figure}[t]
    \centering
    \newcommand*\circled[1]{\tikz[baseline=(char.base)]{
        \node[shape=circle,draw,inner sep=0.5pt] (char) {#1};}}
    \newcommand*\textbox[1]{\tikz[baseline=(char.base)]{
        \node[fill=white,fill opacity=0.9,anchor=south,align=center,text opacity=1] {#1};}}
    \vspace{22pt}
    \begin{overpic}[width=1.0\columnwidth]{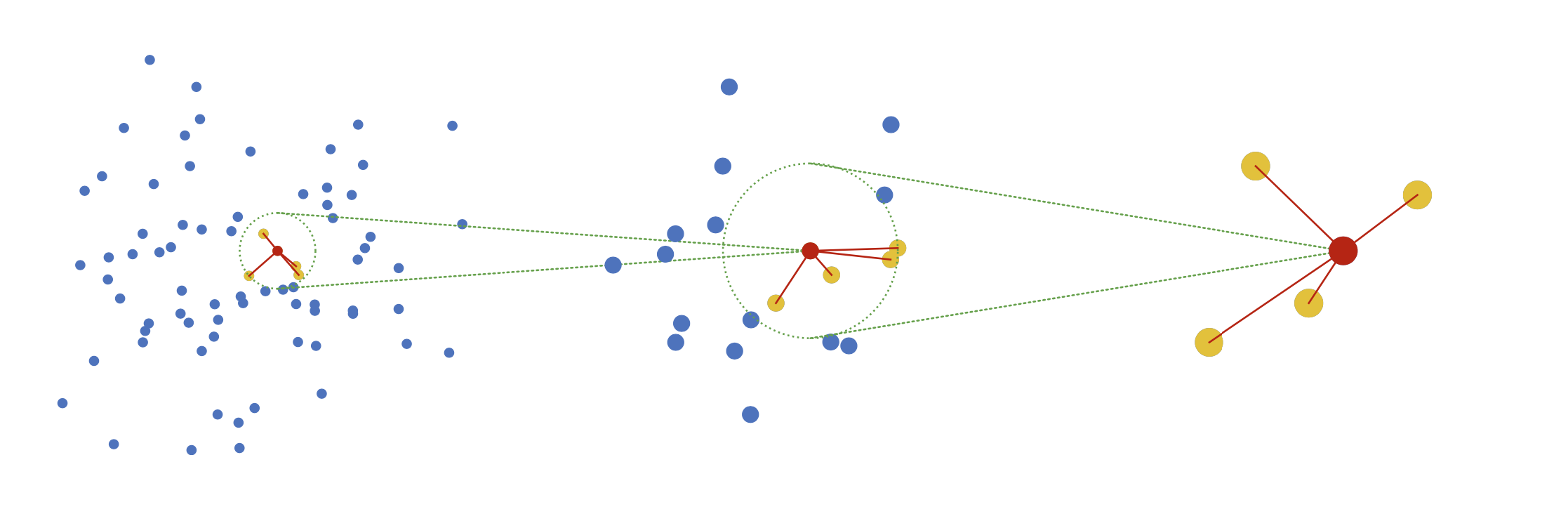}
        \linethickness{1pt}
        \put(27,11){\vector(1,0){15}}
        \put(42,22){\vector(-1,0){15}}
        \put(61,11){\vector(1,0){15}}
        \put(76,22){\vector(-1,0){15}}
        \put(29,12){\tiny \textbox{\circled{1} Coarsen}}
        \put(63,12){\tiny \textbox{\circled{2} Coarsen}}
        \put(63,23){\tiny \textbox{\circled{4} Interpolate}}
        \put(29,23){\tiny \textbox{\circled{6} Interpolate}}
        \put(88,15){\tiny \textbox{\circled{3} Attention}}
        \put(52,18){\tiny \textbox{\circled{5} Attention}}
        \put(15,19.5){\tiny \textbox{\circled{7} Attention}}
        \put(13, 33.5){\parbox{1.2cm}{\centering $0$-th \\ hierarchy}}
        \put(30, 33.5){\Huge \ldots}
        \put(47, 33.5){\parbox{1.2cm}{\centering $h$-th \\ hierarchy}}
        \put(65, 33.5){\Huge \ldots}
        \put(81, 33.5){\parbox{1.2cm}{\centering $H$-th \\ hierarchy}}
    \end{overpic}
    \vspace*{-7mm}
    \caption{The operations of \textbf{one GHA layer} (not a complete network).
    The operation order is shown with circled numbers.
    See \refsec{sec:gha} for a detailed explanation.}
    \label{fig:gha_main}
\end{figure}

Computing full attention on the raw point cloud is in general not feasible, due to the large number of points in a scene.
We propose a global hierarchical attention mechanism, which approximates the full attention while significantly reducing the memory consumption.
The core of GHA is a series of hierarchical levels, augmented with three key ingredients: the \textit{coarsening} operation and the \textit{interpolation} operation that connect tokens between two hierarchies, and the \textit{neighborhood topology} that connects tokens within a single hierarchy (\reffig{fig:gha_main}).

Specifically, starting from $\tilde{Q}^{(0)}$~=~$Q$ (and identically for $K$ and $V$), the coarsening operation
\begin{equation}
\label{eqn:gha_coarsen}
    \tilde{Q}^{(h+1)} = \text{Coarsen} ( \tilde{Q}^{(h)}, \mathcal{T}^{(h)} ) \;
\end{equation}
recursively computes a (down-sampled) representation for each hierarchy level by averaging points via a defined neighborhood topology $\mathcal{T}^{(h)}$ (until all remaining points belong to a single neighborhood), and the interpolation operation
\begin{equation}
\label{eqn:gha_interpolate}
    \tilde{Y}^{(h)} = \text{Interp} ( \tilde{Y}^{(h+1)} ) \;
\end{equation}
expands the coarsened representation to the spatial resolution of the level above (here $Y^{(h)}$ represents any arbitrary embedding).
With these two operation, one can first compute the coarsened $\tilde{Q}^{(h)}$, $\tilde{K}^{(h)}$, $\tilde{V}^{(h)}$ for all hierarchy levels $h$ and approximate the attention weighted output $Z$ via the recursion
\begin{equation}
\label{eqn:gha_recursion}
\begin{aligned}
    \tilde{A}_{ij}^{(h)} &= e^{\tilde{S}_{ij}^{(h)}} = e^{\frac{\tilde{Q}_{i}^{(h)}(\tilde{K}_{i}^{(h)})^T}{\sqrt{d}}} \\
    \tilde{Y_i}^{(h)} &= \sum_{j \in \mathcal{T}_i^{(h)}} \tilde{A}_{ij}^{(h)} \tilde{V_j}^{(h)} + \text{Interp} ( \tilde{Y}^{(h+1)} )_{i} \\
    \tilde{D_i}^{(h)} &= \sum_{j \in \mathcal{T}_i^{(h)}} \tilde{A}_{ij}^{(h)} + \text{Interp} ( \tilde{D}^{(h+1)} )_{i} \\
    Z &\approx \tilde{Z} = \text{diag}(\tilde{D}^{(0)})^{-1} \tilde{Y}^{(0)}.
\end{aligned}
\end{equation}
This procedure is summarized in Algorithm~\ref{alg:gha}.
Note that here, the attention at each level is only computed within the local neighborhood $\mathcal{T}$.
For $N$ points, the memory cost is proportional to $\sum_{h=0}^{H} \frac{1}{k^h} Nk =\frac{1-k^{-(H+1)}}{1-k^{-1}}Nk=\frac{k^{H+1}-1}{k^{H-1}(k-1)}N$ with $H=log_{k}N$ level of hierarchies.
The sum of this series approaches $\frac{k^2}{k-1}N$ with $H\rightarrow\infty$, showing the complexity of GHA is upper bounded by a linear function.
This linear complexity makes it possible to process large-scale point clouds.

The procedure outlined in \refeq{eqn:gha_coarsen}-\ref{eqn:gha_recursion} assigns higher attention weights for closer points (see supplementary for an example).
Thus, GHA natively embodies the inductive bias to focus more on local neighborhoods while retaining the global connectivity.
In this light, local attention as in~\cite{Zhao21ICCV,Park21arXiv} can be viewed as a special case of GHA which truncates all hierarchy below the initial input level.

\begin{algorithm}
\caption{GHA for Point Cloud Analysis}\label{alg:gha}
\begin{algorithmic}
\Require $Q$, $K$, $V \in \mathbb{R}^{N \times d}$
\Ensure $\tilde{Z} \approx \text{softmax}( \frac{QK^{T}}{\sqrt{d}} ) V$
\State $\tilde{Q}^{(0)}, \tilde{K}^{(0)}, \tilde{V}^{(0)} \gets Q, K, V$  \Comment{Coarsening}
\For{$h$ in $(0, \dots, H-1)$}
    \State compute $\tilde{Q}^{(h+1)}$, $\tilde{K}^{(h+1)}$, $\tilde{V}^{(h+1)}$ via \refeq{eqn:gha_coarsen}
\EndFor
\State $\tilde{Y}^{(H)}, \tilde{D}^{(H)} \gets 0, 0$  \Comment{Interpolation}
\For{$h$ in $(H-1, \dots, 0)$}  
    \State compute $\tilde{Y}^{(h)}, \tilde{D}^{(h)}$ via \refeq{eqn:gha_recursion}
\EndFor
\State $\tilde{Z} \gets \text{diag}(\tilde{D}^{(0)})^{-1} \tilde{Y}^{(0)}$
\end{algorithmic}
\end{algorithm}

In order to apply GHA to both point-based and voxel-based methods, we propose two flavors of GHA.
Taking a voxelized point cloud as input, \textbf{Voxel-GHA} uses average pooling as the coarsen operation, unpooling as the interpolation operation, and the local kernel window as the neighborhood topology $\mathcal{T}$.
These operations can be efficiently accomplished using existing sparse convolution libraries~\cite{Choy19CVPR,spconv}.
\textbf{Point-GHA} operates directly on the raw point clouds instead.
It uses \textit{kNN} as the neighborhood topology, and conducts coarsening via farthest point sampling
$\tilde{Q}^{(h+1)} = \text{Sample}_{(h)} ( \frac{1}{K} \sum_{j \in \mathcal{T}_i^{(h)}} \tilde{Q}^{(h)} )$,
and interpolation based on nearest neighbor interpolation
$\tilde{Y}^{(h)} = \text{NN} ( \tilde{Y}^{(h+1)} )$.
Having to compute \textit{kNN} neighborhoods and perform sampling means Point-GHA has some computational overhead compared to its voxel-based counterpart.
In a block composed of multiple GHA layers, the sampling and \textit{kNN} neighborhood computation only needs to be done once and can be shared between all layers, amortizing this overhead. 

\subsection{GHA Block}
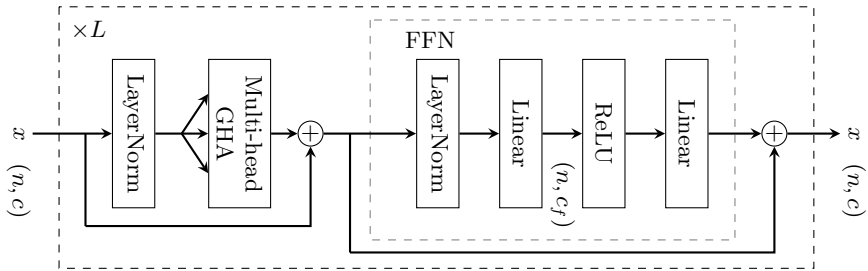
\begin{figure}
    \centering
    \rotatebox{-90}{
    \tikzset{%
        nn-node/.style={text width=8em, minimum width=55pt, text width=50pt, minimum height=16pt,draw, text centered},
        nn-edge/.style={thick},
        nn-arrow/.style={->, >=stealth},
        back group/.style={draw=black!50, dashed, inner xsep=1pt, inner ysep=10pt},
    }
    \begin{tikzpicture}
        \node [] (inp) {$x$};
        \node [right=2pt of inp] (dim1) {$(n,c)$};
        \coordinate[above=20pt of inp] (inpsplit);
        \coordinate[right=35pt of inpsplit] (inpsplitright);
        \node [nn-node, above=10pt of inpsplit] (ln1) {LayerNorm};
        \node [rotate=90, above left=-7pt and 4pt of ln1.west] (l) {$\times L$};
        \coordinate[above=10pt of ln1] (ln1split);
        \node [nn-node, above=20pt of ln1] (gha) {Multi-head GHA};
        \coordinate[left=15pt of gha.south] (ghaleft);
        \coordinate[right=15pt of gha.south] (gharight);
        \node [circle, draw, above=10pt of gha,inner sep=0pt,minimum size=1pt] (skip1) {$+$};
        \coordinate[above=10pt of skip1] (inpsplit2);
        \coordinate[right=45pt of inpsplit2] (inpsplit2right);
        \node [nn-node, above=35pt of skip1] (ln2) {LayerNorm};
        \node [rotate=90, above left=10pt and 1pt of ln2.west] (ffn)  {FFN};
        \node [nn-node, above=15pt of ln2] (linear1) {Linear};
        \node [above right=0pt and 7pt of linear1.north] (dim2) {$(n,c_f)$};
        \node [nn-node, above=15pt of linear1] (relu) {ReLU};
        \node [nn-node, above=15pt of relu] (linear2) {Linear};
        \node [circle, draw, above=20pt of linear2,inner sep=0pt,minimum size=1pt] (skip2) {$+$};
        \node [above=20pt of skip2] (output) {$x$};
        \node [right=2pt of output] (dim3) {$(n,c)$};
        
        \draw[nn-edge, nn-arrow] (inp) -- (ln1);
        \draw[nn-edge, nn-arrow] (ln1) -- (gha);
        \draw[nn-edge, nn-arrow] (ln1split) -- (ghaleft);
        \draw[nn-edge, nn-arrow] (ln1split) -- (gharight);
        \draw[nn-edge, nn-arrow] (gha) -- (skip1);
        \draw[nn-edge] (inpsplit) -- (inpsplitright);
        \draw[nn-edge, nn-arrow] (inpsplitright) |- (skip1.east);
        \draw[nn-edge, nn-arrow] (skip1) -- (ln2);
        \draw[nn-edge, nn-arrow] (ln2) -- (linear1);
        \draw[nn-edge, nn-arrow] (linear1) -- (relu);
        \draw[nn-edge, nn-arrow] (relu) -- (linear2);
        \draw[nn-edge, nn-arrow] (linear2) -- (skip2);
        \draw[nn-edge] (inpsplit2) -- (inpsplit2right);
        \draw[nn-edge, nn-arrow] (inpsplit2right) |- (skip2.east);
        \draw[nn-edge, nn-arrow] (skip2) -- (output);
        
        \begin{scope}[on background layer]
            \node (bk1) [back group] [fit=(ln2) (linear2) (ffn) (dim2)] {};
            \node (bk2) [back group, draw=black, inner xsep=6pt] [fit=(inpsplit) (skip2) (inpsplit2right) (ln2) (ffn)] {};
        \end{scope}
    \end{tikzpicture}}
    \caption{
        In our GHA block the attention and FFN are repeated $L$ times.
        $n$ is the number of points/voxels and $c$/$c_f$ are the feature dimensionalities.
        (Position embeddings are omitted.)
    }
    \label{fig:gha_block}
\end{figure}

Following Vaswani~\etal~\cite{Vaswani17NIPS}, we combine GHA with a feedforward network (FFN) into a GHA layer and stack a total of $L$ layers within a GHA Block as shown in~\reffig{fig:gha_block}.
The input to a block consists of $n$ tokens with a dimensionality of $c$, which we keep throughout the block, apart from the dimensionality $c_f$ that is used between the two linear layers of the FFN.
We apply dropout during training, with probability 0.1 in the attention layer, and 0.3 in the FFN layers.

In order to give the network spatial information, we compute random Fourier mappings~\cite{Tancik20NeurIPS} 
$\gamma (\mathbf{p}) = [cos(2\pi \mathbf{b}_1^T\mathbf{p}), sin(2\pi \mathbf{b}_1^T\mathbf{p}), \cdots, cos(2\pi \mathbf{b}_{m}^T\mathbf{p}), sin(2\pi \mathbf{b}_{m}^T\mathbf{p})]^T$
and use them as positional embedding when computing attention 
($\mathbf{p} \in \mathbb{R}^3$ and $\textbf{b}_i \sim \mathcal{N}(0, 1)$ are sampled at initialization).
Our method uses positional embeddings based on relative positions
$\mathbf{y}_i = \sum_j \mathbf{q}_i^T (\mathbf{k}_j + \gamma (\mathbf{p}_i - \mathbf{p}_j)) \mathbf{v}_j$,
but we also conduct ablation on using absolute positions
$\mathbf{y}_i = \sum_j (\mathbf{q}_i + \gamma (\mathbf{p}_i))^T (\mathbf{k}_j + \gamma (\mathbf{p}_j)) \mathbf{v}_j$.

\section{Evaluation}
\label{sec:evaluation}
Our evaluation can be split into three main parts.
We first evaluate Voxel-GHA for the task of 3D semantic segmentation and secondly for 3D object detection.
In both cases we extend different sparse CNN based approaches with a GHA block and evaluate how this affects their performance.
In the third part we evaluate how Point-GHA can serve as a replacement for vanilla global attention in the recent 3DETR object detection approach~\cite{Misra21ICCV}.

\subsection{Voxel-GHA: 3D Semantic Segmentation}

\textbf{Dataset.} We conduct our experiment on the challenging ScanNet dataset~\cite{Dai17CVPR}.
It contains 3D reconstructions of indoor rooms collected using RGB-D scanners and points are labeled with 20 semantic classes.
We follow the official training, validation, and test split (1201, 312, and 100 scans).

\noindent \textbf{Setting.}
We use two MinkUNet~\cite{Choy19CVPR} architectures as baselines, the smaller ME-UNet18 and the bigger ME-UNet34, and test them on two voxel sizes, 2\,cm and~5\,cm.
Both networks are based on ResNet encoders (18 and 34 layers, respectively) and a symmetric decoder.
We insert $L=6$ GHA layers ($c=256, c_f=128$, 8 heads) to the decoder branch after the convolution block at $2\times$ voxel stride (at $4\times$ voxel stride for 2\,cm experiments, in order to roughly match the number of voxels in the attention layer).
On average, this resulted in 5-12k voxels per frame as input for the attention, a number for which computing multiple layers of full global attention is generally infeasible.
We use a $3\times3\times3$ local window for the neighborhood in GHA (which we apply for all of our experiments).
See supplementary for details on the training.

\begin{table}[t]
    \centering
    \setlength{\tabcolsep}{4pt}
    \captionof{table}{3D semantic segmentation performance (mIoU) on ScanNet validation set using original or cleaned labels.}
    \label{tab:scannet_seg}
    \begin{tabular}{ l cc c cc}
        \toprule
        & \multicolumn{2}{c}{Original} && \multicolumn{2}{c}{Clean~\cite{Ye21ICCV}} \\
        \cline{2-3} \cline{5-6}
        Method & 5\,cm & 2\,cm && 5\,cm & 2\,cm \\
        \midrule
        ME-UNet18 & 67.8 & 72.6 && 68.4 & 73.2 \\
        \quad + local attention & 69.0 & \textbf{73.3} && 69.6 & \textbf{73.8} \\
        \quad + local attention ($L=8$) & 68.8 & -- && 69.3 & -- \\
        \quad + GHA & \textbf{69.6} & 72.8 && \textbf{70.2} & 73.4 \\
        \midrule
        ME-UNet34 & 68.3 & 72.9 && 69.0 & 73.7 \\
        \quad + local attention & 68.6 & \textbf{73.6} && 69.0 & \textbf{74.6} \\
        \quad + local attention ($L=8$) & 69.1 & -- && 69.7 & -- \\
        \quad + GHA & \textbf{70.0} & 73.5 && \textbf{70.3} & 74.1\\
        \bottomrule
    \end{tabular}
\end{table}

\noindent \textbf{Main results.}
In \reftab{tab:scannet_seg} we present the mean intersection-over-union (mIoU) between the predicted and labeled masks on the ScanNet validation set (see supplementary for per-class performance).
In addition to the official annotation, we also report an evaluation done using the cleaned validation labels from~\cite{Ye21ICCV}.
Compared to the CNN only baseline, GHA gives a clear improvement across all network models and voxel sizes.
The performance gap is especially visible for the coarser 5\,cm voxels, where GHA improves the baseline by 1.7\%~mIoU.
At finer voxel sizes, however, we see that the improvement of GHA is small, only 0.2\%~mIoU, for the weaker ME-UNet18 backbone.
We speculate that, at this fine scale, a smaller backbone may not be strong enough to extract rich features that can support the global fitting of GHA.
The results using cleaned validation labels are in general higher, but the trend matches that observed using original labels, suggesting that the networks do not have special behavior with respect to labeling noise.

\begin{figure}[t]
    \newcommand{\figrowentry}[3]{%
        \begin{tikzpicture}[every node/.style={inner sep=0,outer sep=0}]\node [anchor=south west] (image) at (0,0){\includegraphics[width=0.25\textwidth,trim={#2},clip]{fig/qual_scannet/#1.pdf}};#3\end{tikzpicture}%
    }%
    \newcommand{\figrowscannetframed}[3]{%
        \frame{\figrowentry{#1_rgb}{#2}{#3}}%
        \frame{\figrowentry{#1_conv}{#2}{#3}}%
        \frame{\figrowentry{#1_gha}{#2}{#3}}%
        \frame{\figrowentry{#1_gt}{#2}{#3}}%
    }%
    \newcommand{\figrowscannet}[3]{%
        \figrowentry{#1_rgb}{#2}{#3}%
        \figrowentry{#1_conv}{#2}{#3}%
        \figrowentry{#1_gha}{#2}{#3}%
        \figrowentry{#1_gt}{#2}{#3}%
    }%
    \newcommand{\overbox}[4]{%
        \draw[thick,gs_red] (#1,#2) rectangle (#3,#4);%
    }%
    \centering
    \begin{tabularx}{\textwidth}{YYYY}%
    RGB & MinkUNet & GHA  & Ground truth\\%
    \end{tabularx}%
    \hspace{0.2mm}\figrowscannet{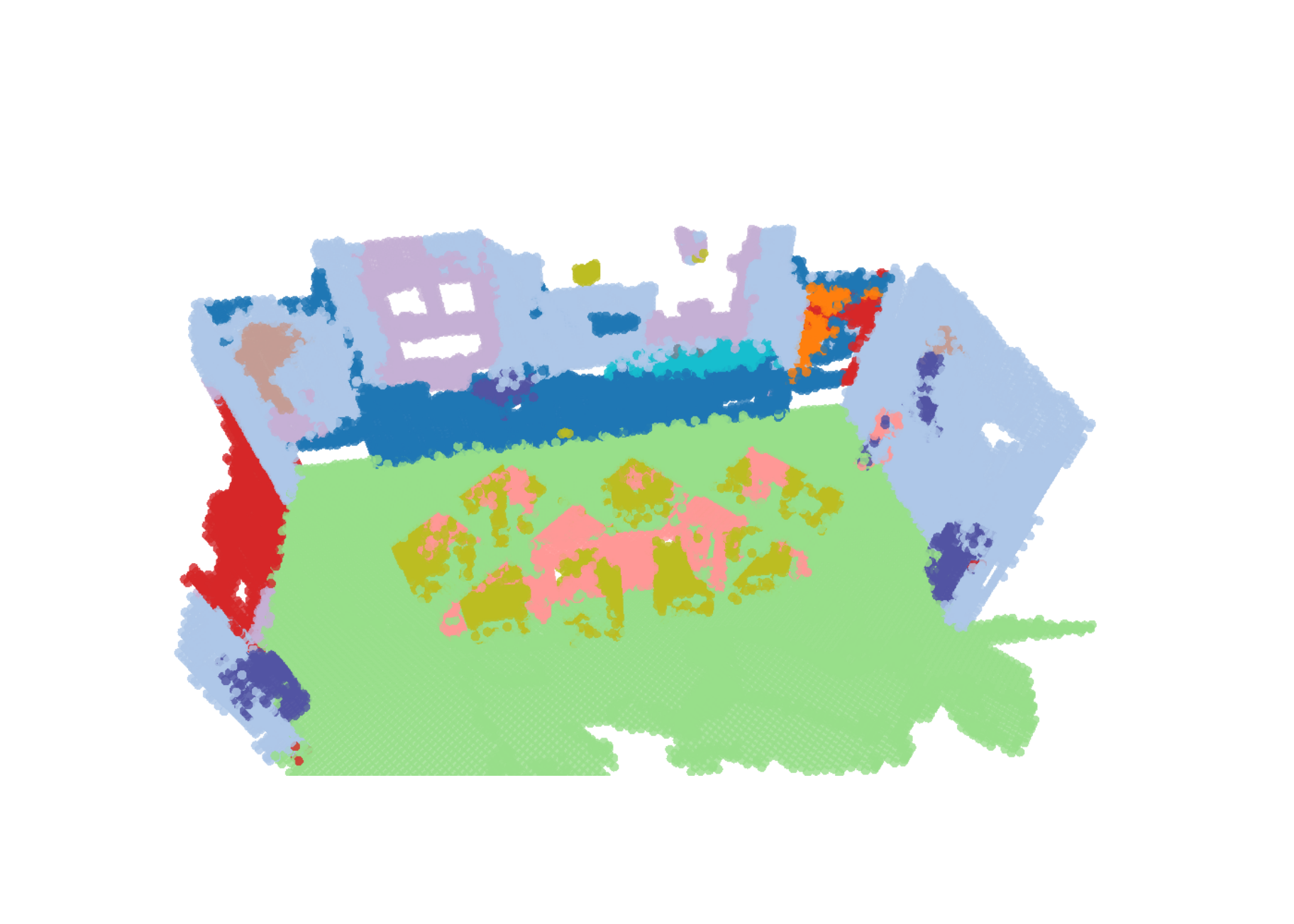}{2.3cm 2.3cm 2.3cm 2.9cm}{%
        \overbox{0.12}{0.88}{0.72}{1.52}%
        \overbox{1.83}{1,0}{2.34}{1.6}%
    }%
    \hspace{0.2mm}\figrowscannet{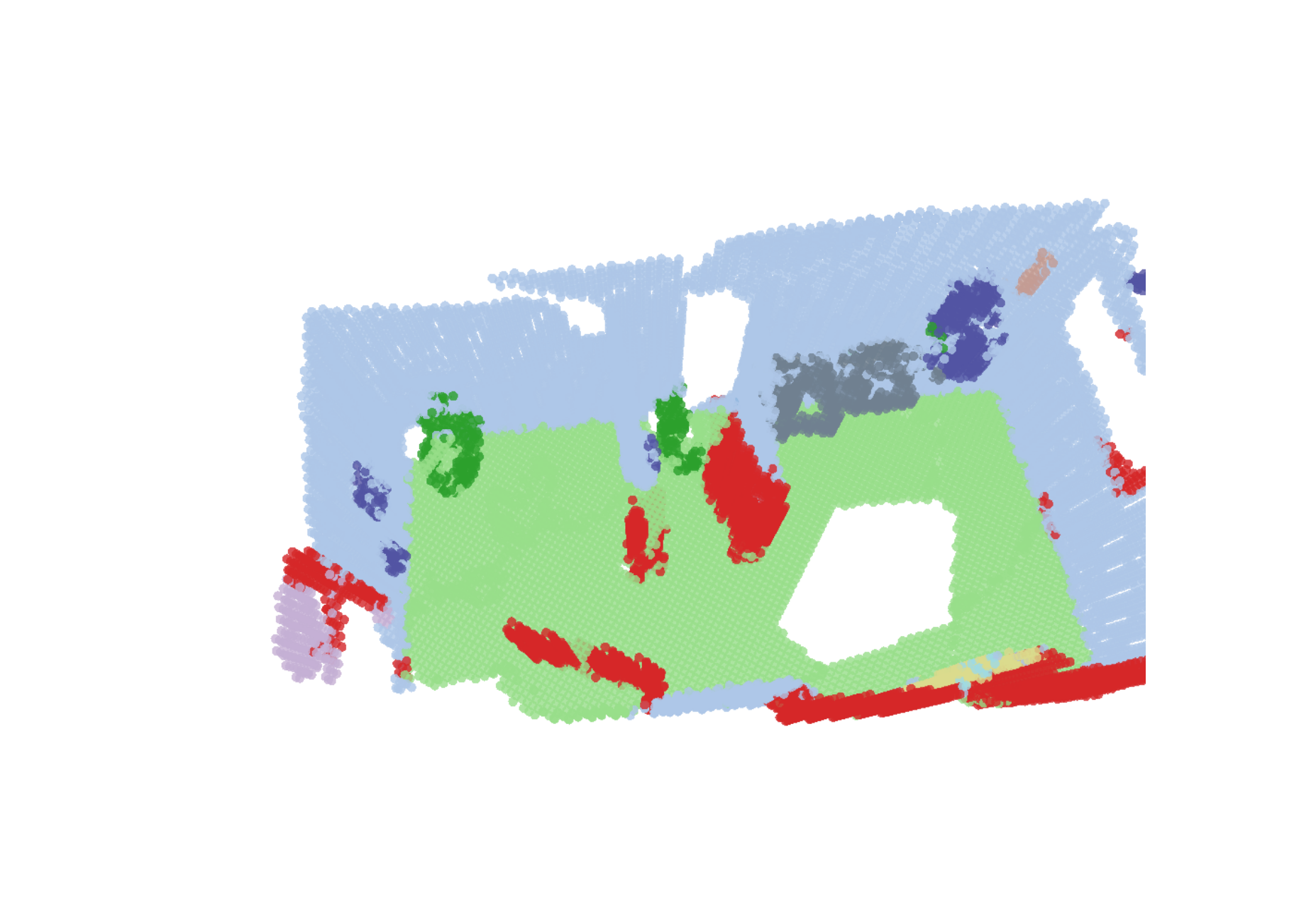}{3.5cm 2.3cm 2.3cm 2.3cm}{%
        \overbox{0.05}{0.2}{0.55}{0.8}%
    }%
    \hspace{0.2mm}\figrowscannet{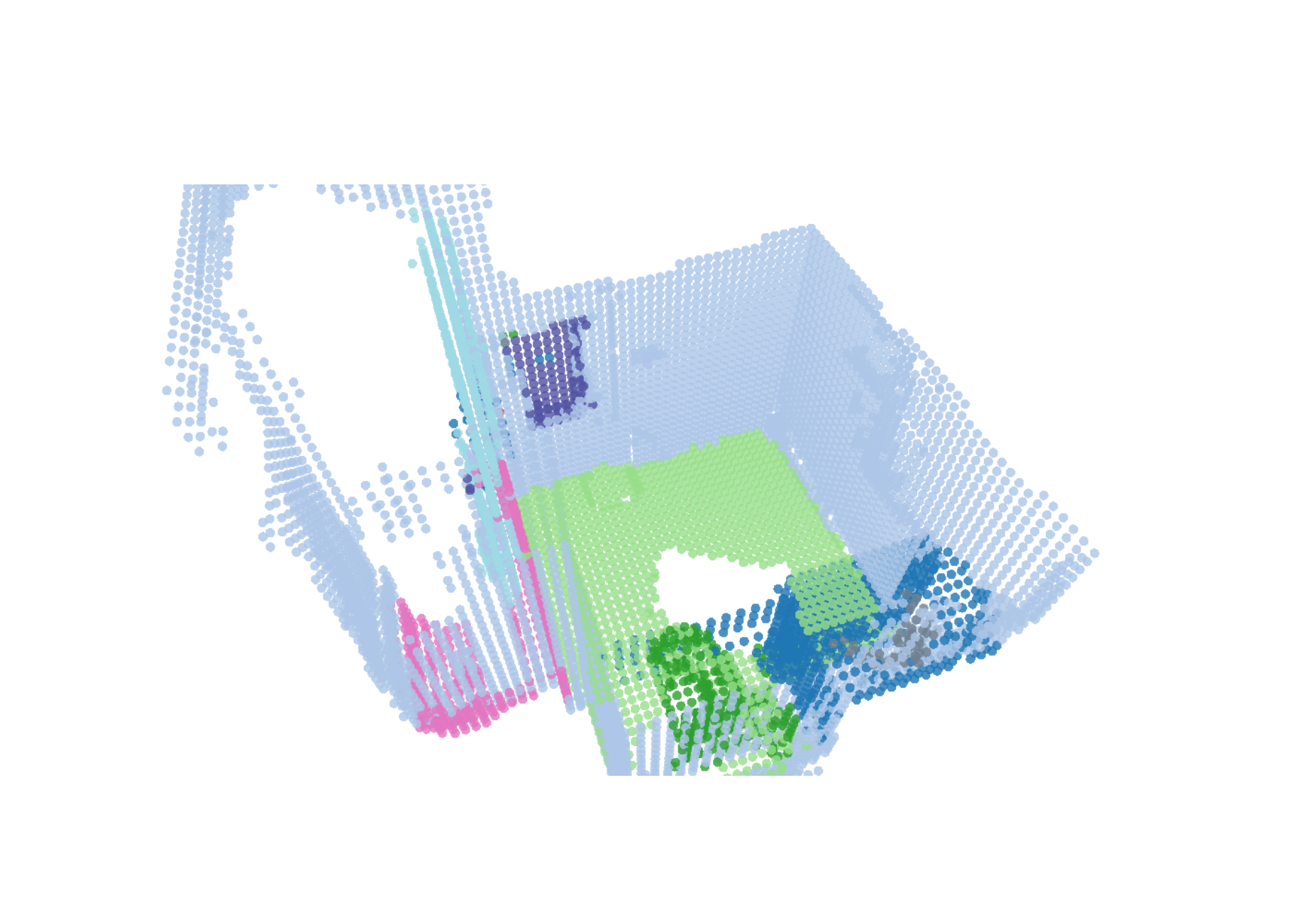}{2.3cm 2.3cm 2.3cm 2.3cm}{
        \overbox{1.35}{0.85}{2.1}{1.72}%
    }%
    \hspace{0.2mm}\figrowscannet{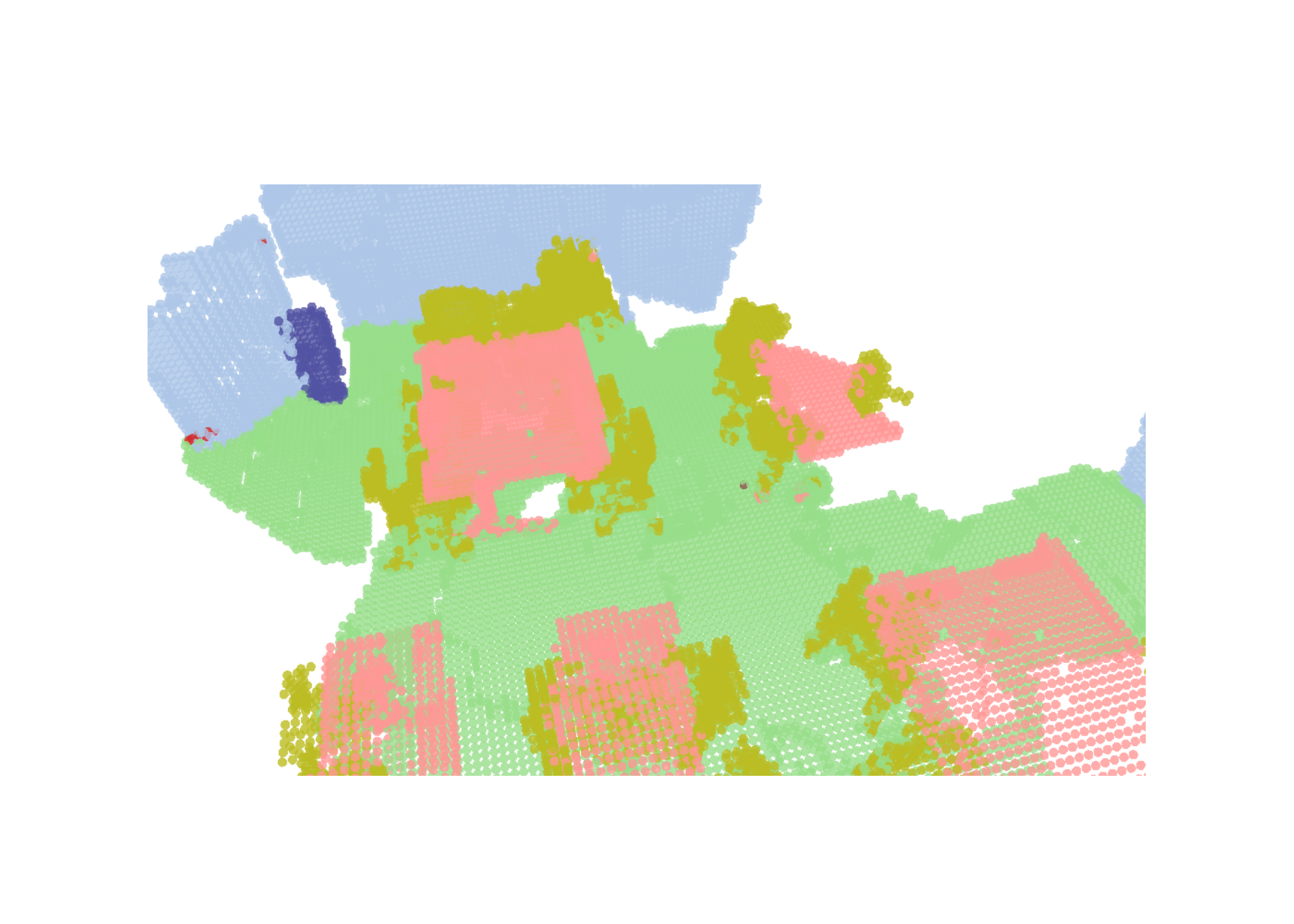}{2.0cm 2.3cm 2.5cm 2.3cm}{
        \overbox{0.0}{0.85}{0.6}{1.75}%
    }%
    \hspace{0.2mm}\figrowscannet{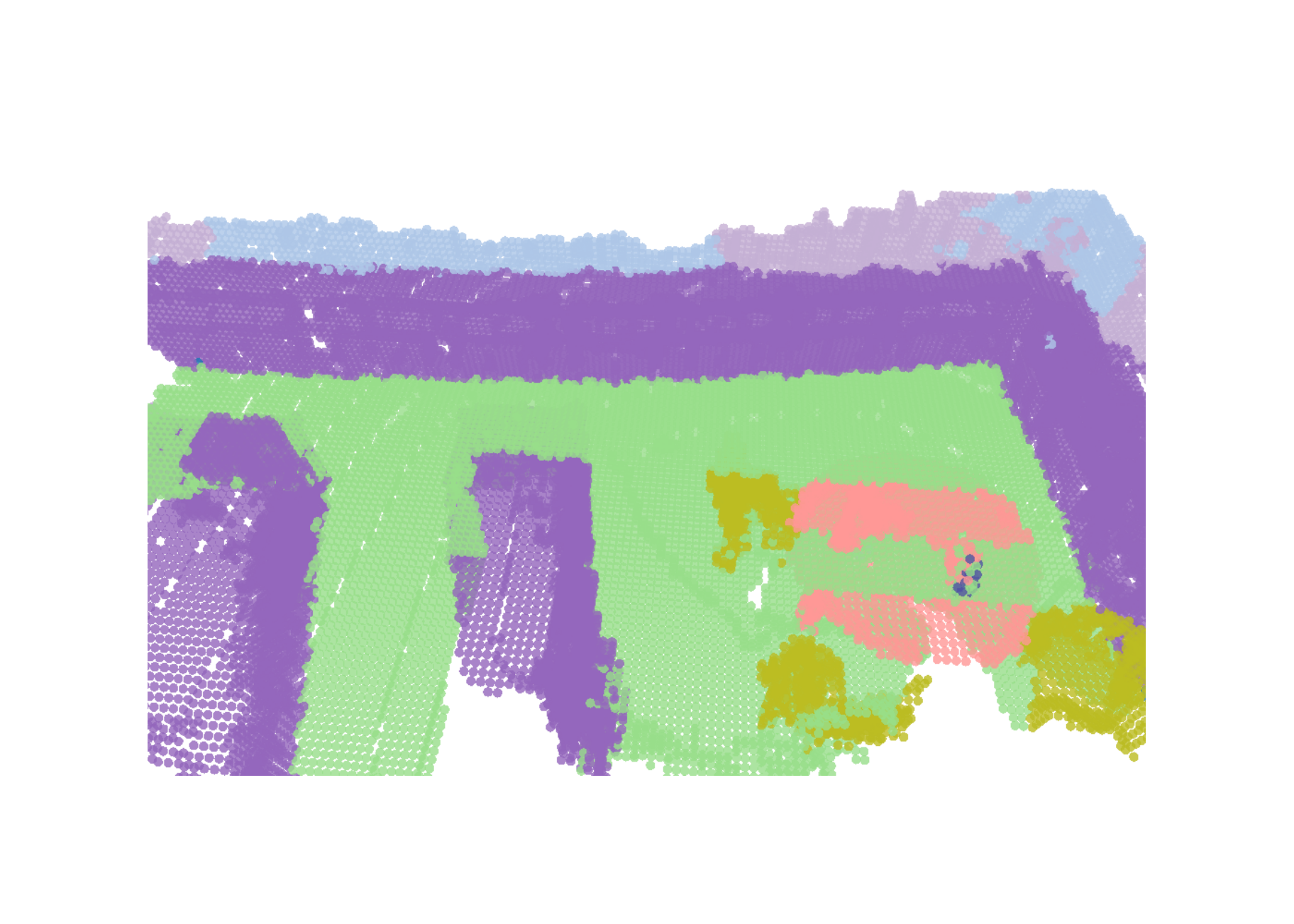}{2.3cm 2.3cm 2.3cm 2.3cm}{
        \overbox{0.0}{1.35}{2.5}{1.8}%
    }%
    
    \scannetcoltable
    
    \caption{
        Qualitative semantic segmentation results on the ScanNet validation set.
        Predictions with GHA show clear improvement in the marked regions, where semantic labels are ambiguous judging from local geometry and color information alone.
        This improvement highlights the benefit of long-range information.
    }
    \label{fig:scannet_seg_qualiative}
\end{figure}
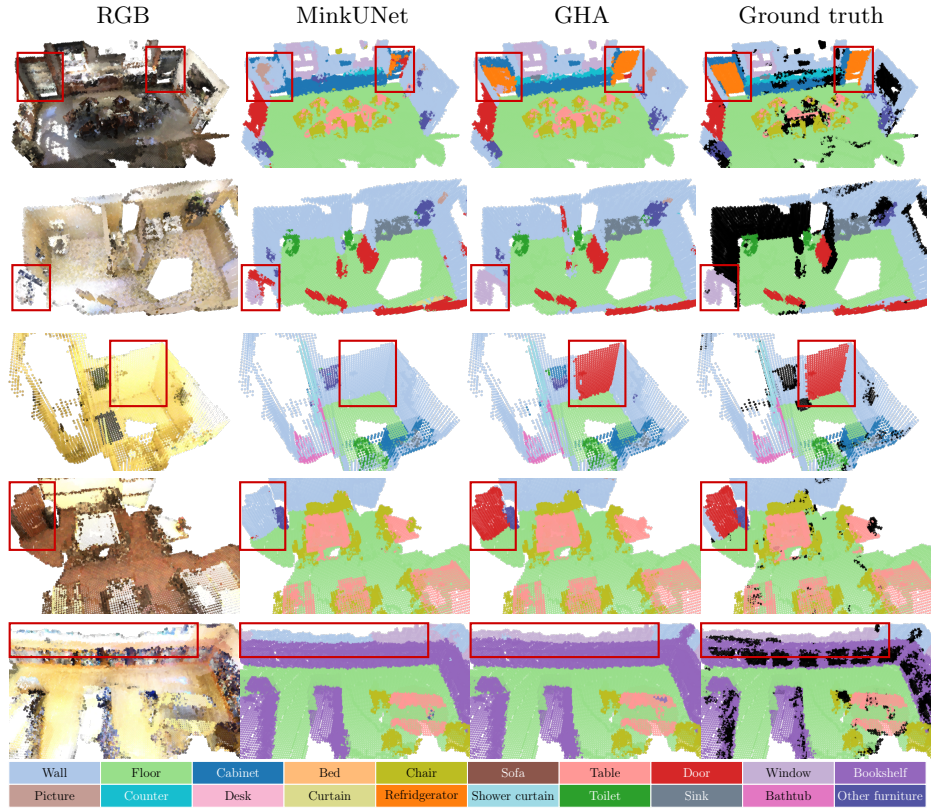

\begin{figure}
    \centering
    \includegraphics{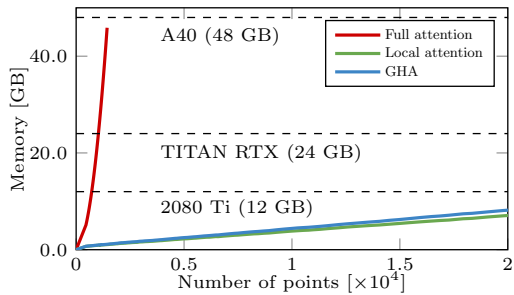}
    \captionof{figure}{Training memory measured for 6 attention layers.}
    \label{fig:attn_memory}
\end{figure}

In addition, we conduct ablation studies by replacing GHA with local attention.
Since local attention consumes less memory than GHA, to have a fair comparison, we also experiment with a bigger model composed of eight layers, which surpasses memory consumption of GHA.
With 5\,cm voxels, using the same memory, GHA outperforms local attention by 0.8\%~mIoU.
This shows the benefit of having a global field of view.
However, with 2\,cm voxels, local attention outperforms GHA.
This may suggest that, depending on the resolution and the capacity of the feature extractor, limiting field of view can be beneficial to the overall performance, and prompts further architecture level investigation.
\reffig{fig:scannet_seg_qualiative} shows some qualitative results.
The predictions of GHA are significantly better on the regions where the semantic labels are ambiguous based on local geometry and color information alone, showing the advantage of including non-local information for inference.
\reffig{fig:attn_memory} shows the memory consumption of one forward and backward pass for different attention types.
GHA has linear space complexity, similar to that of local attention, while having a global field of view.
The memory consumption of full attention is prohibitively expensive.
Training the ME-18 network (5\,cm) takes 34 hours for GHA and 27.5 hours for local attention, both using a single A40 GPU.
For inference on ScanNet, GHA averaged 5.6 and 3.7 FPS for 5 and 2\,cm voxels (local attention has 12.9 and 6.5 FPS).
The inference speed can be improved with a cuda-optimized indexing implementation and in-place aggregation.

\begin{figure}[t]
    \centering
    \begin{subfigure}[b]{0.50\textwidth}
    \resizebox{\linewidth}{!}{\includegraphics{fig/attn_hist.tikz}}
    \end{subfigure}%
    \hspace{0.019\textwidth}%
    \begin{subfigure}[b]{0.47\textwidth}%
    \begin{overpic}[angle=270,origin=c,trim={0 0.5cm 1cm 0.5cm},clip,width=\columnwidth]{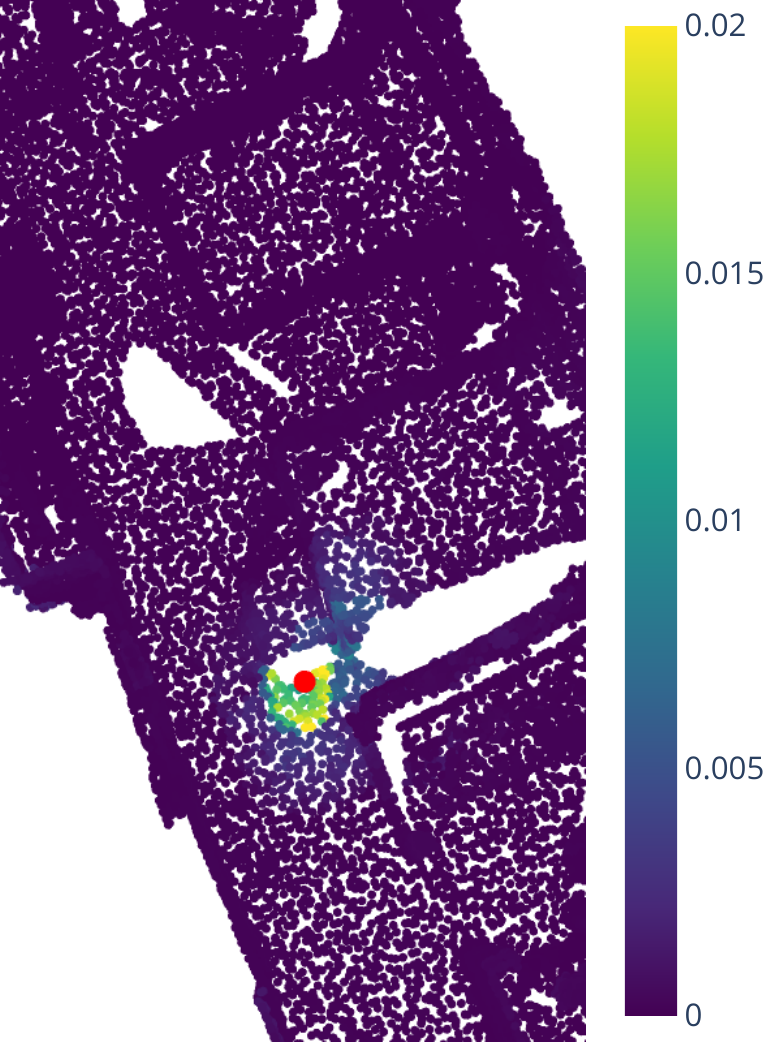}
        \put(0,10){\scriptsize 0.00}
        \put(91,10){\scriptsize 0.02}
        \put(44.5,10){\scriptsize 0.01}
    \end{overpic}
    \end{subfigure}%
    
    \caption{
    \textbf{Left:} The distribution of attention distances for both the GHA and local attention.
    Local attention is incapable of attending to voxels beyond a fixed threshold, whereas GHA can and does indeed attend globally.
    Notice, however, the logarithmic y-axis.
    \textbf{Right:} The attention heatmap of a \textbf{randomly initialized} GHA layer (source point marked with \raisebox{1pt}{\protect\tikz\protect\draw[red,fill=red] (0,0) circle (.34ex);}).
    Closer points tend to have higher weights.
    Both figures show the locality bias of GHA.
    }
    \label{fig:attn_dist}
\end{figure}

\reffig{fig:attn_dist}\,(Left) shows the distribution of attention at different distances.
While local attention only attends within a local neighborhood, GHA has a global receptive field, and can attend to all regions in the scene.
In addition, GHA has the inductive bias to place an emphasis on nearby points, leading to the skewed pattern in the attention histogram.
\reffig{fig:attn_dist}\,(Right) shows the attention of a randomly initialized GHA layer as a heatmap.
Its inductive bias can clearly be seen, as close points tend to receive higher weights, even without being trained.

\noindent \textbf{Additional experiments.}
We conduct experiments evaluating several design choices. These experiments are done on ScanNet using the ME-UNet34 model with 5\,cm voxels, and the results are presented in~\reftab{tab:gha_additional}.

\textit{Number of layers.}
Experiments on different numbers of layers show that, with sufficient network depth, GHA outperforms the baseline or the local attention variants.
We attribute this superior performance to its global receptive field, which channels information beyond a local neighborhood.

\textit{Positional embedding.}
We experiment with different types of positional embedding, including the learned embedding via a two-layer MLP.
Results show that random Fourier mapping with relative positions gives the best performance while requiring no additional parameters.

\textit{Attention type.}
In addition to the standard scaled dot-product attention~\cite{Vaswani17NIPS}, cosine-similarity-based attention~\cite{Park21arXiv} and full vector attention~\cite{Zhao20CVPR} have also been used in transformer architectures for point clouds.
The full vector attention gives a 1.2\%~mIoU performance increase, at the cost of increased memory requirements and parameter counts (training batch size needs to be halved).
We use the regular dot-product attention to stay consistent with most existing research.

\textit{Local window vs. kNN neighbors.}
Our method uses $3\times3\times3$ local windows as the neighborhood $\mathcal{T}$ in~\refeq{eqn:gha_recursion}.
As an alternative, we can use a \textit{kNN}-neighborhood ($k=27$), resulting in on-par performance (69.9\%~mIoU).
We opt for the faster local window-based neighborhood, which in addition has the advantage of being robust against varying point densities.

\begin{table}[t]
    \centering
    \setlength{\tabcolsep}{4pt}
    \caption{Influence of different factors on the ScanNet validation set performance.}

    \resizebox{0.187\textwidth}{!}{
    \begin{tabular}{ l c }
        \toprule
        \# Layers & mIoU \\
        \midrule
        2 & 68.3 \\
        4 & 68.6 \\
        6 & 70.0 \\
        8 & 69.5 \\
        \bottomrule
    \end{tabular}}
    \quad
    \resizebox{0.308\textwidth}{!}{
    \begin{tabular}{ l c c }
        \toprule
        Embedding & Relative & mIoU \\
        \midrule
        None &    & 69.0 \\
        Learnable & \ding{55} & 69.3 \\
        Fourier & \ding{55} & 69.4 \\
        Fourier & \ding{51} & 70.0 \\
        \bottomrule
    \end{tabular}}
    \quad
    \begin{tabular}{ l c }
        \toprule
        Attention type & mIoU \\
        \midrule
        Cosine~\cite{Park21arXiv} & 68.8 \\
        Vector~\cite{Zhao21ICCV} & 71.2 \\
        Dot-product~\cite{Vaswani17NIPS} & 70.0 \\
        \bottomrule
    \end{tabular}
    \label{tab:gha_additional}
\end{table}

\subsection{Voxel-GHA: 3D Object Detection}

\noindent \textbf{Dataset.}
We conduct our experiments using the \textit{KITTI}~\cite{Geiger13IJRR} and the \textit{nuScenes} dataset~\cite{Caesar20CVPR}, both are collected using cars equipped with LiDAR sensors driving on the city streets.
The KITTI dataset contains 7,481 LiDAR scans for training and 7,518 for testing. We follow prior work~\cite{Zhou17CVPR} and divide the training split into 3,712 samples for training and 3,769 for validation.
The nuScenes dataset is bigger and more challenging, containing 1,000 sequences (380,000 scans), divided into 700/150/150 sequences for training, validation, and testing.
The KITTI dataset is annotated with bounding boxes for three classes, whereas the nuScenes dataset is annotated with ten classes.

\begin{table}[t]
\centering
\parbox{0.66\textwidth}{
    \centering
    \setlength{\tabcolsep}{1pt}
    \caption{
        3D object detection performance (mAP) on the KITTI validation set for (E)asy, (M)oderate, and (H)ard objects. \textsuperscript{\ding{61}}:~numbers taken from~\cite{mmdet3d2020}.
    }
    \resizebox{0.66\textwidth}{!}{
    \begin{tabularx}{\columnwidth}{l YYY cYYY cYYY}
        \toprule
        \multirow[b]{2}{*}{Method} & \multicolumn{3}{c}{Car} && \multicolumn{3}{c}{Cyclist} && \multicolumn{3}{c}{Pedestrian}\\
        \cmidrule{2-4} \cmidrule{6-8} \cmidrule{10-12}
        & E & M & H && E & M & H && E & M & H\\
        \midrule
        VoxelNet~\cite{Zhou17CVPR} & 82.0 & 65.5 & 62.9 && 67.2 & 47.7 & 45.1 && 57.9 & 53.4 & 48.9 \\
        SECOND~\cite{Yan18Sensors} & 87.4 & 76.5 & 69.1 && -- & -- & -- && -- & -- & -- \\
        F-PointNet~\cite{Qi18CVPR} & 83.8 & 70.9 & 63.7 && -- & -- & -- && -- & -- & -- \\
        PointRCNN~\cite{Shi19CVPR} & 89.1 & 78.7 & 78.2 && \textbf{93.5} & \textbf{74.2} & \textbf{70.7} && 65.8 & 59.6 & 52.8 \\
        MVF\textsuperscript{\ding{61}}~\cite{Zhou19CORL}  & 88.6 & 77.7 & 75.1 && 80.9 & 61.9 & 59.5 && 61.3 & 55.7 & 50.9 \\
        Part-$A^2$~\cite{Shi20PAMI} & 89.5 & 79.5 & 78.5 && 88.3 & 73.1 & 70.2 && \textbf{70.4} & \textbf{63.9} & \textbf{57.5} \\
        PV-RCNN~\cite{Shi20CVPR} & \textbf{92.6} & \textbf{84.8} & \textbf{82.7} && -- & -- & -- && -- & -- & -- \\
        3DSSD~\cite{Yang20CVPR} & 89.7 & 79.5 & 78.7 && -- & -- & -- && -- & -- & -- \\
        \midrule
        SECOND  & 88.6 & 77.6 & \textbf{74.8} && 80.4 & 66.7 & 63.1 && 61.5 & 55.5& 50.1\\
        \quad + LA              & \textbf{89.0} & \textbf{77.8} & 74.5                            && 76.8 & 61.7 & 59.7                            && 60.2 & 53.2 & 49.2 \\
        \quad + LA ($L=6$)      & 88.0 & 77.5 & 74.4                            && 77.7 & 64.2 & 60.1                            && 63.5 & 55.8 & 51.2 \\
        \quad + GHA             & 88.8 & \textbf{77.8} & 74.6          && \textbf{81.7} & \textbf{67.8} & \textbf{63.9} && \textbf{64.0} & \textbf{56.1} & \textbf{51.2}\\
        \bottomrule 
    \end{tabularx}}
    \label{tab:kitti_dec}
}
\hspace{0.01\textwidth}
\parbox{0.3\textwidth}{
    \centering
    \setlength{\tabcolsep}{4pt}
    \caption{
        3D object detection performance on nuScenes validation set. 
    }
    \resizebox{0.305\textwidth}{!}{
    \begin{tabular}{ l c c }
        \toprule
        Method & mAP & NDS \\
        \midrule
        PointPillars~\cite{Lang19CVPR} & 28.2 & 46.8 \\
        3DSSD~\cite{Yang20CVPR} & 42.6 & 56.4 \\
        CenterPoint~\cite{Yin21CVPR} & \textbf{56.4} & \textbf{64.8} \\
        SASA~\cite{Chen22AAAI} & 45.0 & 61.0 \\ 
        \midrule
        CenterPoint (10\,cm) & 56.2 & 64.4 \\
        \quad + LA & \textbf{56.8} & \textbf{64.8} \\
        \quad + LA ($L=6$) & \textbf{56.8} & 64.7 \\
        \quad + GHA & 56.7 & 64.7 \\
        \midrule
        CenterPoint (7.5\,cm) & 57.3 & 65.2 \\
        \quad + LA & 57.7 & 65.6 \\
        \quad + LA ($L=6$) & \multicolumn{2}{c}{OOM} \\
        \quad + GHA  & \textbf{57.9} & \textbf{65.7} \\        
        \bottomrule
    \end{tabular}}
    \label{tab:nusc_det}
}
\end{table}

\noindent \textbf{Setting.}
We use implementations from the MMDetection3D library~\cite{mmdet3d2020} for the next set of our experiments.
For the KITTI dataset, we experiment with augmenting the SECOND detector~\cite{Yan18Sensors}  with GHA.
The SECOND detector uses a sparse voxel-based backbone to encode point cloud features, which are then flattened into a dense 2D feature map and passed into a standard detection head for bounding box regression.
We add $L=4$ GHA layers ($c=64, c_f=128$, 4 heads) at the end of the sparse voxel encoder and leave the rest of the network unchanged (the dimension of GHA layers is chosen to match the encoder output).

For the nuScenes dataset, we experiment with the state-of-the-art CenterPoint detector~\cite{Yin21CVPR} which combines a VoxelNet~\cite{Zhou17CVPR} backbone and a center-based object detection head.
Similar to the KITTI experiment, we add $L=4$ GHA layers ($c=128, c_f=256$, 8 heads) at the end of the backbone.
We experiment with two voxel sizes, 10\,cm and 7.5\,cm.
To reduce the training time, we load a pre-trained model from~\cite{mmdet3d2020}, and fine-tune the network end-to-end for three epochs (the pre-trained model was trained for 20 epochs).

\noindent \textbf{Results.}
In~\reftab{tab:kitti_dec} we report the performance of the retrained SECOND detector with and without GHA, along with several other methods for comparison, and in~\reftab{tab:nusc_det} we report detection mAP and NDS (nuScenes detection score) on nuScene dataset.
In all experiment cases, GHA matches or outperforms the baseline results.
On the KITTI dataset, the benefit is especially visible on the smaller cyclist and pedestrian classes (1.1\% and 0.6\%~mAP increase respectively for the moderate setting).
On the nuScenes dataset, fine-tuning with GHA for only 3 epochs consistently brings a 0.5\%~mAP and 0.3\%~NDS improvement for both voxel sizes.
See supplementary for qualitative results.

\subsection{Point-GHA: 3D Object Detection}

\noindent \textbf{Dataset.} To evaluate point-based GHA for object detection, we experiment on ScanNet detection dataset \cite{Dai17CVPR}, which contains 1,513 indoor scans with annotated bounding boxes, split into 1,201 scenes for training and 312 for validation.
Following prior works~\cite{Qi19ICCV,Cheng21CVPR,Cheng20ECCV,Xie21ICCV}, we report the mean average precision, specifically mAP$_{25}$ and mAP$_{50}$, computed at a 0.25 and 0.5 IoU threshold respectively.

\noindent \textbf{Setting.}
We experiment with the 3DETR network proposed by Misra~\textit{et al.}~\cite{Misra21ICCV}.
3DETR follows a transformer encoder-decoder architecture.
From the input point cloud, 3DETR first samples 2048 points and extracts their neighboring features via a single set aggregation operation, as introduced by Qi~\etal in their PointNet++~\cite{Qi17NIPS}.
These points are then processed with a transformer encoder composed of multiple self-attention layers.
During the decoding stage, 3DETR samples a set of query points and projects them into the embedding space (\textit{object queries}), and through a stack of self and cross-attention, decodes each of the object queries into bounding boxes.

To evaluate Point-GHA, we replace 3DETR's encoder self-attention with Point-GHA ($k=6$ for the \textit{kNN} neighborhood), leaving all other network components unchanged.
We refer to this network as 3DETR-GHA.
In addition, Misra~\etal experimented with masking of attention between points further than a threshold, and obtained improved results.
This version, called 3DETR-m, also includes an intermediate pooling layer.
Since GHA inherently has the inductive bias to focus on local neighborhoods, we do not need to apply distance-based masking.
Rather, we experiment with adding the intermediate pooling layer, and refer to this version as 3DETR-GHA-p.
Finally, with the benefit of the reduced memory consumption of GHA, we experiment with a bigger model by including more points (4096) into the encoder.

\begin{table}[t]
\parbox{.53\linewidth}{

    \centering
    \setlength{\tabcolsep}{4pt}
    \caption{3D object detection performance on the ScanNet validation set.}
    \begin{tabularx}{\linewidth}{ X c c }
        \toprule
        Method & mAP$_{25}$ & mAP$_{50}$ \\
        \midrule
        VoteNet~\cite{Qi19ICCV}  & 60.4 & 37.5 \\
        MLCVNet~\cite{Xie20CVPR}  & 64.7 & 42.1 \\
        BRNet~\cite{Cheng21CVPR}    & 66.1 & 50.9 \\
        H3DNet~\cite{Cheng20ECCV}   & 67.2 & 48.1 \\
        VENet~\cite{Xie21ICCV}    & 67.7 & -- \\
        GroupFree3D~\cite{Liu21ICCV} & \textbf{69.1} & \textbf{52.8} \\
        \midrule
        3DETR~\cite{Misra21ICCV}    & 62.7 & 37.5 \\
        3DETR-m~\cite{Misra21ICCV}  & \textbf{65.0} & \textbf{47.0} \\
        \midrule
        GHA-3DETR & 64.3 & 45.6 \\
        GHA-3DETR-p & 65.4 & 46.1 \\
        GHA-3DETR-p\, (4096) & \textbf{67.1} & \textbf{48.5} \\
        \bottomrule 
    \end{tabularx}
    \label{tab:point_gha_results}
}
\hspace{0.04\linewidth}
\parbox{.43\linewidth}{
    \centering
    \setlength{\tabcolsep}{4pt}
    \caption{3DETR performance with different encoder self-attention or \textit{kNN} neighborhood sizes.}
    \resizebox{0.38\textwidth}{!}{
    \begin{tabularx}{\linewidth}{X c c}
        \toprule
        Attention & mAP$_{25}$ & mAP$_{50}$ \\
        \midrule
        Regular & 62.7 & 37.5 \\
        Local & 63.1 & 43.9 \\
        BigBird~\cite{Zaheer20NeurIPS} & 63.4 & 43.4 \\
        \midrule
        GHA & \textbf{64.3} & \textbf{45.6} \\
        \bottomrule 
    \end{tabularx}}%
    \newline%
    \vspace*{1pt}%
    \newline%
    \resizebox{0.38\textwidth}{!}{%
    \begin{tabularx}{\linewidth}{X c c}%
        \toprule
        \textit{kNN} size & mAP$_{25}$ & mAP$_{50}$ \\
        \midrule
        $k=2$ & 64.1 & 42.9 \\
        $k=4$ & 64.1 & 43.6 \\
        $k=6$ & \textbf{64.3} & \textbf{45.6} \\
        $k=8$ & 63.4 & 43.2 \\
        $k=10$ & 62.6 & 44.3 \\
        $k=20$ & 61.7 & 41.1 \\
        \bottomrule 
    \end{tabularx}}
    \label{tab:point_gha_ablate}
}
\end{table}

\noindent \textbf{Results.} 
In~\reftab{tab:point_gha_results} we present the main experiment result.
Compared to the 3DETR baseline, our GHA-3DETR obtains an improvement of 1.6\% on mAP$_{25}$ and 8.1\% on mAP$_{50}$.
Compared to the masking 3DETR-m, our GHA-3DETR-p achieves similar performance.
When including more points, GHA-3DETR-p gives an mAP$_{25}$ and mAP$_{50}$ improvement of 2.1\% and 1.5\% respectively, coming closer to state-of-the-art performance.

In~\reftab{tab:point_gha_ablate} we present an ablation study, where we replace the self-attention in 3DETR with different attention mechanisms.
Compared to the regular attention (used in 3DETR), local attention achieved improved results, especially for mAP$_{50}$, despite not having a global field of view.
This demonstrates the value of focusing on local neighbors.
GHA outperforms local attention, thanks to the additional access to global information acquired from its hierarchical design.
We additionally implement a variant of BigBird~\cite{Zaheer20NeurIPS} to work with point clouds (it was originally proposed for language tasks), which augments local attention with several tokens that have global connectivity (\textit{global attention}), and randomly activate a small number of non-local attentions (\textit{random sparse attention}).
GHA outperforms our BigBird implementation.
Even though both methods have global connectivity and have linear memory complexity, GHA additionally embodies the inductive bias to focus on local information, to which we attribute its better performance.
In~\reftab{tab:point_gha_ablate} we also present a study on the effect of the \textit{kNN} neighborhood size, and the result shows that GHA is robust to a wide range \textit{k}, consistently outperforming its counterpart using regular attention.

\section{Conclusion}
In this paper we have proposed Global Hierarchical Attention (GHA) for point cloud analysis.
Its inductive bias towards nearby points, its global field of view, as well as its linear space complexity, make it a promising mechanism for processing large point clouds.
Extensive experiments on both detection and segmentation show that adding GHA to an existing network gives consistent improvements over the baseline network.
Moreover, ablation studies replacing GHA with other types of attention have demonstrated its advantage.

The overall positive results point to future research directions.
The GHA mechanism can in principle be applied to image-domain tasks as well, which we have not experimented with so far.
In addition, it would be interesting to explore new architectures with GHA, beyond plugging it into existing networks.

\PAR{Acknowledgements.}
This project was funded by the BMBF project 6GEM (16KISK036K) and the ERC Consolidator Grant DeeVise (ERC-2017-COG-773161). We thank Jonas Schult, Markus Knoche, Ali Athar, and Christian Schmidt for helpful discussions.

\newpage

\section{Supplementary Material}
We provides an illustration of GHA using an example (\refsec{sec:gha_example}), training details for all experiments (\refsec{sec:training_settings}), per-class semantic segmentation performance on ScanNet (\refsec{sec:voxel_gha_sem_seg}), and qualitative detection results (\refsec{sec:qual_results}) of GHA-augmented networks.

\subsection{GHA with an example}
\label{sec:gha_example}

\begin{figure}
    \centering
    \includegraphics[width=0.9\columnwidth]{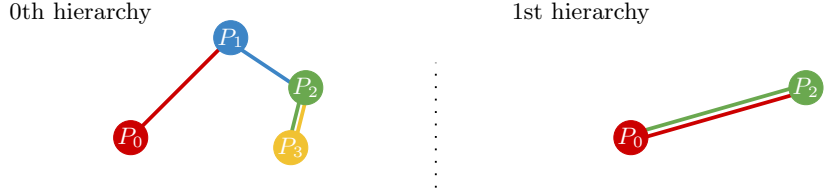}
    \caption{An example of four points with two hierarchy levels. The \textit{kNN} neighborhood ($k=2$) for each point is marked with colored edges.}
    \label{fig:illustration}
\end{figure}

Consider the example shown in~\reffig{fig:illustration}, where there are four points.
We use \textit{kNN} neighborhood with $k=2$, and farthest point sampling with a factor of two, resulting in two hierarchies in total.
For this configuration, the GHA output for the point $P_0$ is given by
\begin{equation}
\label{eqn:gha_example}
\begin{aligned}
    \tilde{Z}_0 &= \frac{1}{\tilde{D}_0}( {\underbrace{\vphantom{\frac{0}{0}} w_0^{(0)} V_0 + w_1^{(0)} V_1}_{\text{from $0$th hierarchy}}} + {\underbrace{w_0^{(1)} \frac{V_0 + V_1}{2} + w_2^{(1)} \frac{V_2 + V_3}{2}}_{\text{from $1$st hierarchy}}} )\\
    &= \frac{1}{\tilde{D}_0}(  (w_0^{(0)} + \frac{w_0^{(1)}}{2}) V_0 + (w_1^{(0)} + \frac{w_0^{(1)}}{2} ) V_1 + \frac{w_2^{(1)}}{2} V_2 + \frac{w_2^{(1)}}{2} V_3 )
\end{aligned}
\end{equation}
with attention weights $w$ and normalization factor $\tilde{D}_1$
\begin{equation}
\label{eqn:gha_example_supp}
\begin{aligned}
    \tilde{D}_0 &= w_0^{(0)} + w_1^{(0)} + w_0^{(1)} + w_2^{(1)} \\
    w_0^{(0)} &= e^{\frac{Q_{0}K_0^{T}}{\sqrt{d}}}, \quad w_1^{(0)} = e^{\frac{Q_{0}K_1^{T}}{\sqrt{d}}} \\
    w_0^{(1)} &= e^{ \frac{ (Q_{0} + Q_{1})(K_{0} + K_{1})^T }{4\sqrt{d}} } \\
    w_2^{(1)} &= e^{ \frac{ (Q_{0} + Q_{1})(K_{2} + K_{3})^T }{4\sqrt{d}} }.
\end{aligned}
\end{equation}
The effect of GHA is clearly illustrated: neighboring points ($P_0$, $P_1$) are assigned with higher weights,
while the interaction with far points ($P_2$, $P_3$) are only approximated with a single attention weight, reducing the overall memory footprint.

\subsection{Training settings}
\label{sec:training_settings}

\textbf{ScanNet semantic segmentation.}
We train all networks using the AdamW optimizer~\cite{Loshchilov19ICLR} and a one-cycle policy~\cite{Smith19AIMLMDOA} for 600 epochs with batch size 8.
For the MinkowskiEngine baseline, we use a learning rate of 1$e$-2 with 0.01 weight decay, and for the GHA-augmented network, we use a learning rate of 1$e$-3 with 0.05 weight decay, which performed better in the respective settings.
For augmentation, we use random rotation, scaling, and random cropping~\cite{Zhang21ICCV}.

\textbf{KITTI object detection.}
We use the original setting from~\cite{mmdet3d2020}, namely, we train the network for 40 epochs, with batch size 6, AdamW optimizer~\cite{Loshchilov19ICLR}, 0.018 learning rate, and a one cycle learning rate policy~\cite{Smith19AIMLMDOA}.

\textbf{nuScenes object detection.}
We use a batch size of 8 for 10\,cm voxels, and 6 for 7.5\,cm voxels, the AdamW optimizer~\cite{Loshchilov19ICLR}, and a learning rate of $1e-5$ with a one-cycle policy~\cite{Smith19AIMLMDOA} with 0.05 weight decay.

\textbf{ScanNet object detection (Point-GHA).}
Following~\cite{Misra21ICCV}, we train our network for 1080 epochs, with the AdamW optimizer~\cite{Loshchilov19ICLR}, a cosine learning rate schedule, with a final learning rate of 1$e$-6, weight decay 0.1, and 0.1 $L2$-norm gradient clipping.

\subsection{Per-class performance on ScanNet semantic segmentation}
\label{sec:voxel_gha_sem_seg}

\reftab{tab:scannet_per_class} shows the per-class performance of the~2\,cm experiments, along with several state-of-the-art methods.
Compared to the baseline, GHA performs particularly well on the ``fridge" class, which has very few training samples, but under-performs on the ``shower" class.
This behavior may be caused by the global field of view of GHA, which brought in supportive (in the ``fridge" case) or distracting (in the ``shower" case) long range information.

\begin{table*}
\centering
\caption{Per-class semantic segmentation score on ScanNet validation set. \textsuperscript{\ding{61}}:~numbers taken from~\cite{Nekrasov213DV}.}
\label{tab:scannet_per_class}
\resizebox{\textwidth}{!}{%
\setlength{\tabcolsep}{1.8pt}
\begin{tabular}{rc cccccccccccccccccccc}
        \toprule
        Method & mIoU & wall & floor & cabinet & bed & chair & sofa & table & door & wndw & books. & pic & counter & desk & curtain & fridge & shower & toilet & sink & bathtub & otherf.\\
        \midrule
        KPConv\textsuperscript{\ding{61}}\,\cite{Thomas19ICCV} & 69.3 & 82.4 & 94.4 & 64.5 & 79.2 & 88.5 & 77.2 & 73.0 & 60.5 & 59.1 & \textbf{79.8} & 28.4 & 59.9 & \textbf{63.7} & 71.6 & \textbf{53.1} & 54.1 & 91.5 & 63.3 & 86.0 & \textbf{56.4} \\
        MinkNet\textsuperscript{\ding{61}}\,\cite{Choy19CVPR} & 72.4 & \textbf{85.6} & \textbf{96.5} & \textbf{64.7} & \textbf{82.1} & \textbf{91.0} & \textbf{84.4}  & \textbf{74.5} & \textbf{65.0} & \textbf{62.8} & 79.5 & \textbf{32.4} & \textbf{64.4} & \textbf{63.7} & \textbf{75.5} & 51.6  &  \textbf{69.0} & \textbf{93.0}  & \textbf{67.6} & \textbf{87.6}  & 56.3 \\
        BPNet~\cite{Hu21CVPR} & \textbf{73.9} & -- & -- & -- & -- & -- & -- & -- & -- & -- & -- & -- & -- & -- & -- & -- & -- & --  & -- & -- & -- \\
        VMNet~\cite{Hu21ICCV} & 73.3 & -- & -- & -- & -- & -- & -- & -- & -- & -- & -- & -- & -- & -- & -- & -- & -- & --  & -- & -- & -- \\
        \midrule
        ME-UNet18 & 72.6 & 85.1 & 95.2 & 65.2 & 79.0 & \textbf{91.7} & 84.1 & 73.8 & 63.2 & 64.6 & \textbf{80.4} & 32.2 & 62.1 & 63.9 & 75.2 & 55.9 & \textbf{70.8} & 92.8 & 67.3 & 86.4 & 62.3 \\
        + local   & \textbf{73.3} & 85.6 & \textbf{95.4} & 66.7 & 79.6 & 91.2 & \textbf{83.7} & 74.8 & \textbf{66.1} & \textbf{65.4} & 80.2 & \textbf{33.5} & \textbf{65.8} & 66.3 & \textbf{77.0} & 55.4 & 69.3 & \textbf{93.6} & \textbf{66.9} & 86.1 & \textbf{62.8} \\
        + GHA     & 72.8 & \textbf{85.9} & 95.1 & \textbf{69.4} & \textbf{79.9} & 91.0 & 83.6 & 75.3 & 63.6 & 65.3 & \textbf{80.4} & 30.8 & 61.1 & \textbf{67.6} & 76.0 & \textbf{62.0} & 60.6 & 92.6 & \textbf{66.9} & \textbf{87.0} & 61.1 \\
        \midrule
        ME-UNet34 & 72.9 & 85.1 & 95.2 & 65.6 & 78.6 & 91.5 & 84.1 & 73.9 & 62.2 & 64.7 & 79.4 & 33.0 & \textbf{64.9} & 65.1 & 76.3 & 60.0 & 71.7 & 93.3 & 65.7 & 86.4 & 61.1 \\
        + local   & \textbf{73.6} & \textbf{86.1} & \textbf{95.4} & 66.4 & 79.1 & \textbf{91.7} & 83.7 & 74.3 & \textbf{64.5} & \textbf{64.9} & 80.3 & \textbf{33.2} & 63.6 & 64.2 & \textbf{77.4} & 62.6 & \textbf{73.8} & \textbf{93.7} & 68.3 & 86.1 & \textbf{63.7} \\
        + GHA     & 73.5 & 85.7 & 95.3 & \textbf{67.0} & \textbf{79.9} & 91.0 & \textbf{85.1} & \textbf{77.5} & 61.7 & \textbf{64.9} & \textbf{81.8} & 31.0 & 63.7 & \textbf{68.9} & 76.0 & \textbf{65.3} & 63.2 & 93.0 & \textbf{68.7} & \textbf{87.6} & 61.9 \\
        \bottomrule
\end{tabular}%
}
\end{table*}

\subsection{Qualitative results for object detection}
\label{sec:qual_results}

\begin{figure}
    \includegraphics[width=1.0\textwidth,trim={1cm 9cm 2cm 4cm},clip]{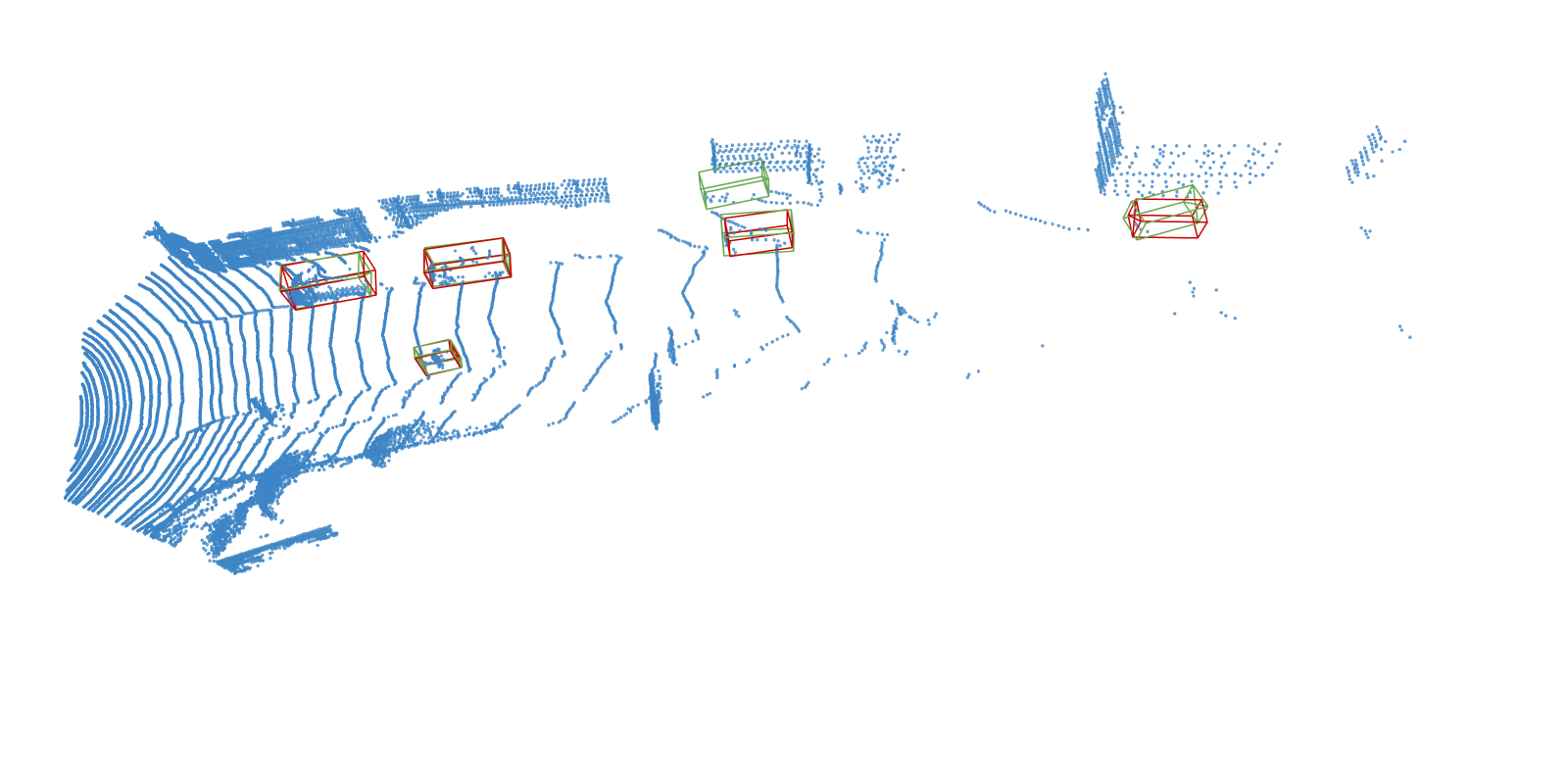}\\%
    \includegraphics[width=1.0\textwidth,trim={2cm 10cm 4cm 0cm},clip]{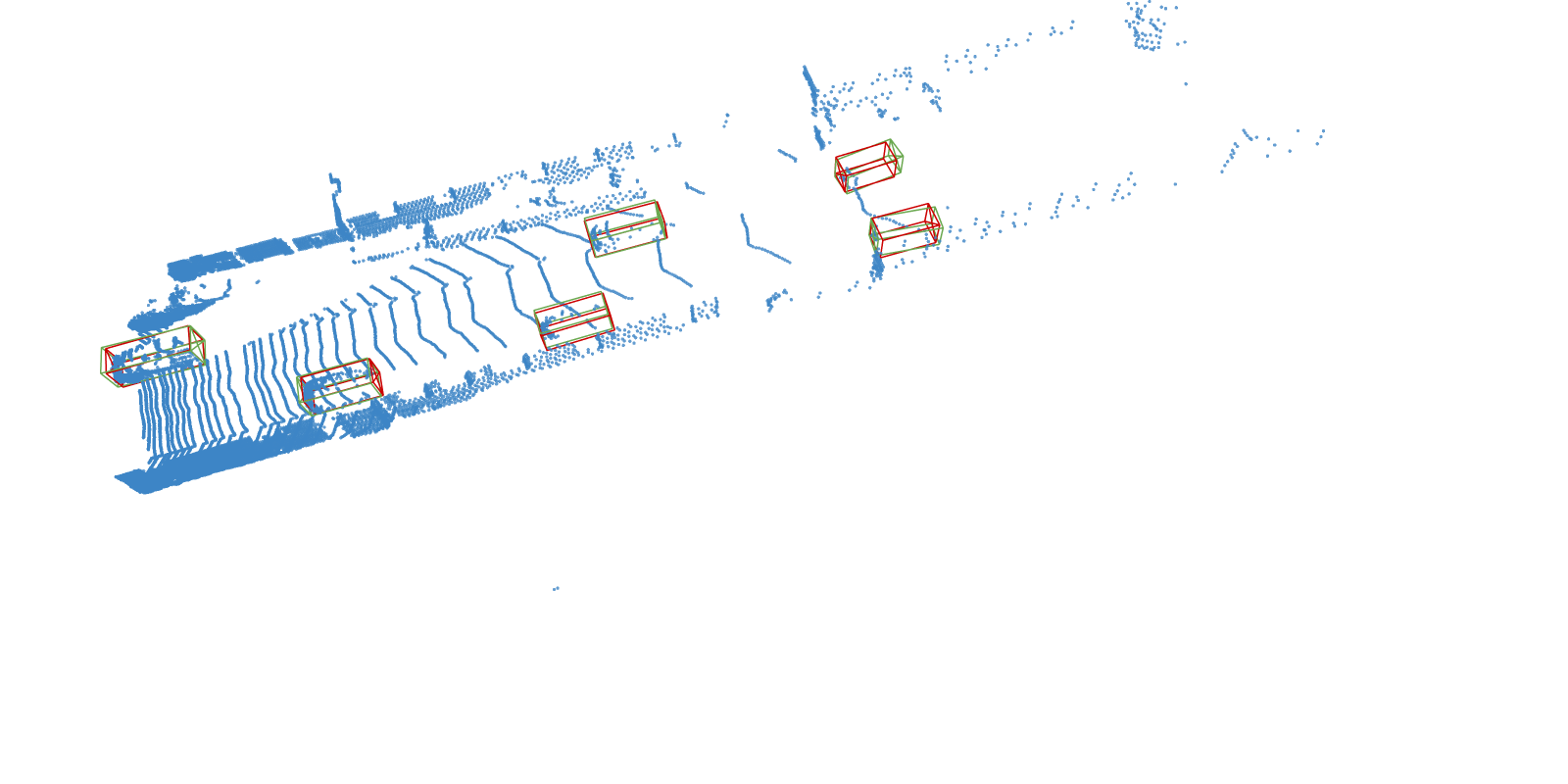}\\%
    \includegraphics[width=1.0\textwidth,trim={5cm 5cm 4cm 0cm},clip]{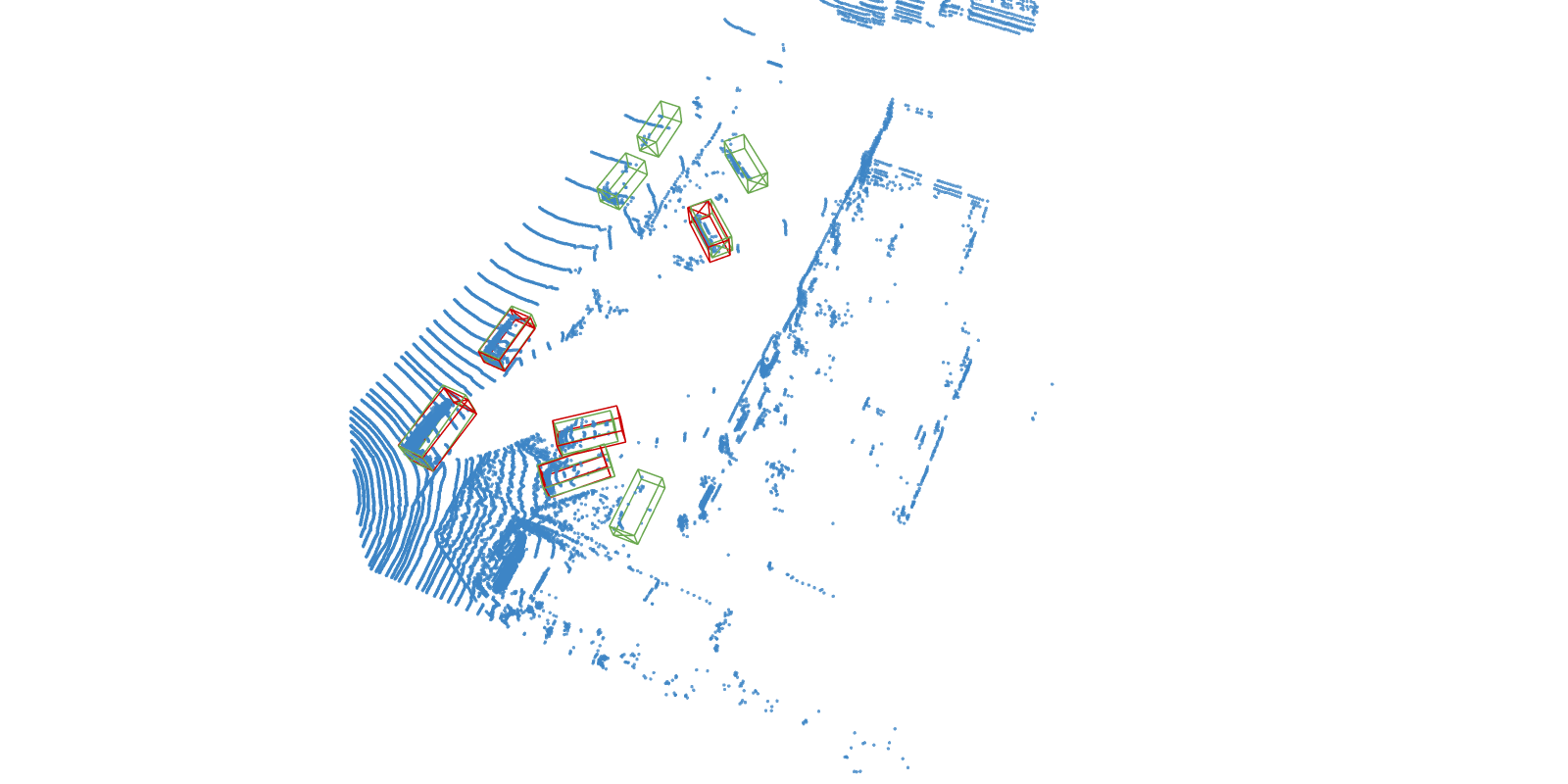}
    \caption{
        Qualitative detection results of SECOND~\cite{Yan18Sensors} with GHA on the KITTI~\cite{Geiger12CVPR} validation set.
        {\color{gs_green} Predictions} and {\color{gs_red} groundtruth} are shown as {\color{gs_green} green} and {\color{gs_red} red} bounding boxes.
    }
    \label{fig:kitti_qual_begin}
\end{figure}

\begin{figure}
    \includegraphics[width=1.0\textwidth,trim={0cm 5cm 4cm 1.5cm},clip]{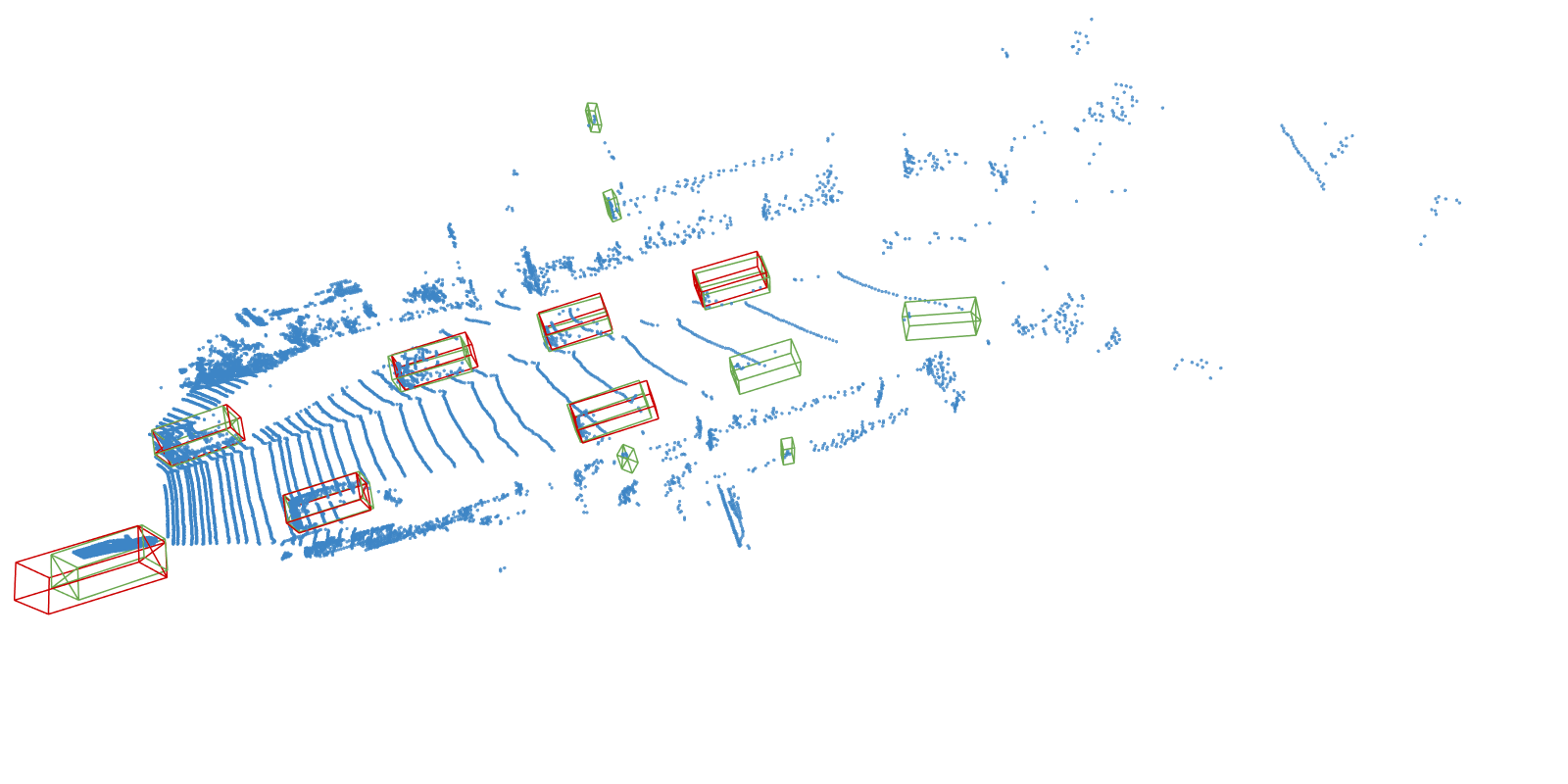}\\%
    \includegraphics[width=1.0\textwidth,trim={4cm 2cm 3cm 2cm},clip]{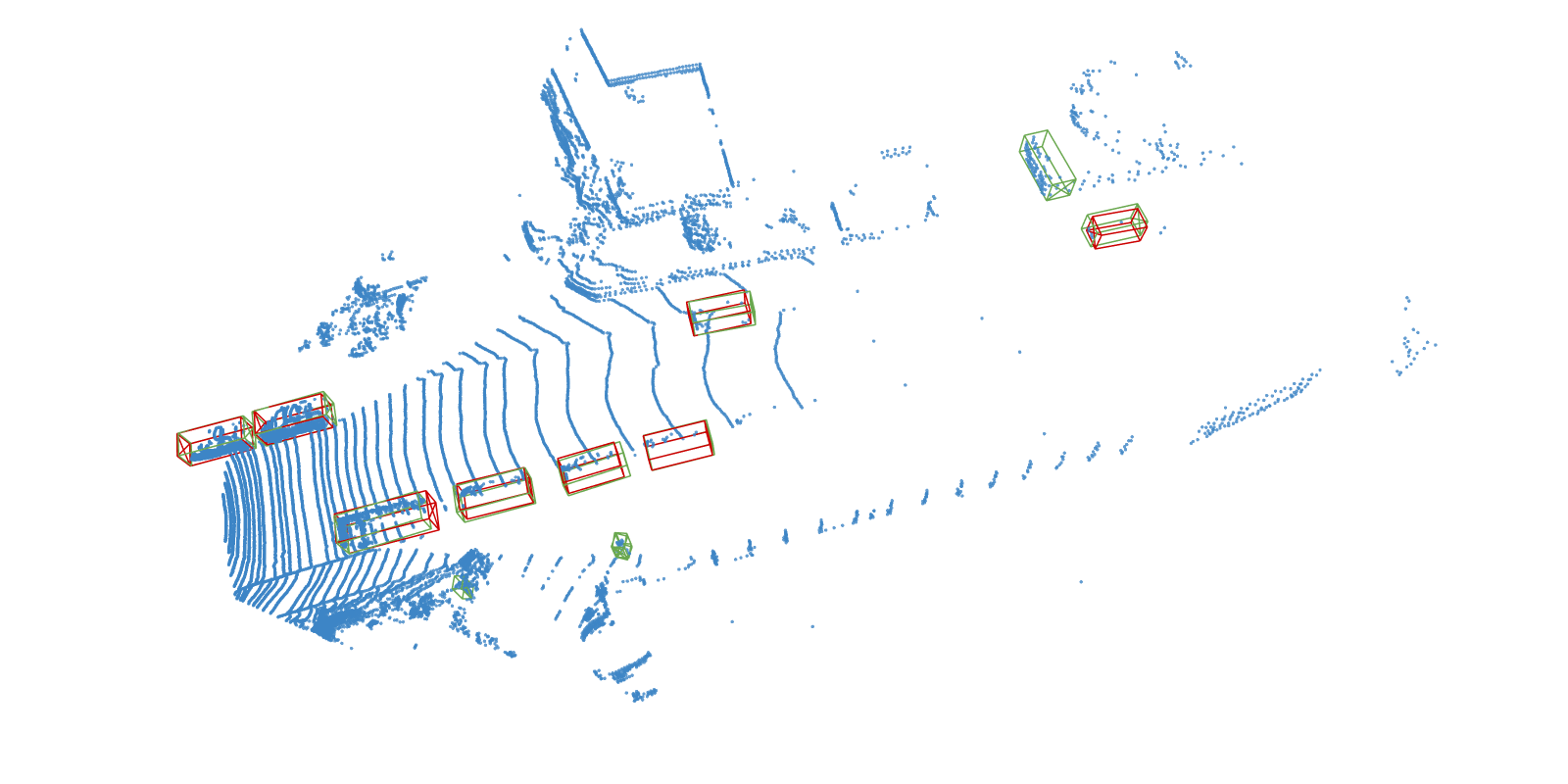}\\%
    \includegraphics[width=1.0\textwidth]{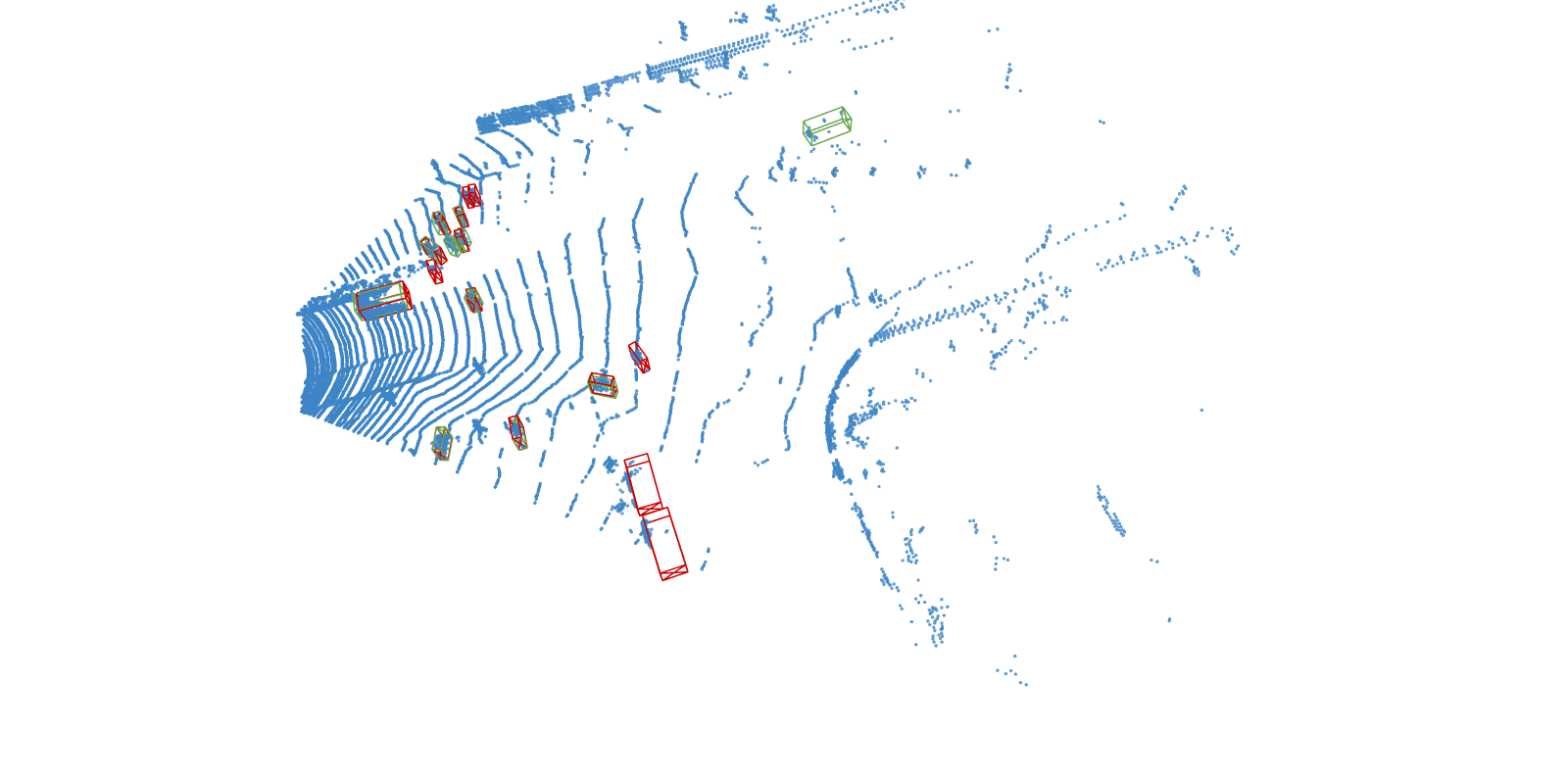}
    \caption{
        Qualitative detection results of SECOND~\cite{Yan18Sensors} with GHA on the KITTI~\cite{Geiger12CVPR} validation set.
        {\color{gs_green} Predictions} and {\color{gs_red} groundtruth} are shown as {\color{gs_green} green} and {\color{gs_red} red} bounding boxes.
    }
    \label{fig:kitti_qual_end}
\end{figure}

\begin{figure}
    \includegraphics[width=1.0\textwidth,trim={0cm 0cm 0cm 3cm},clip]{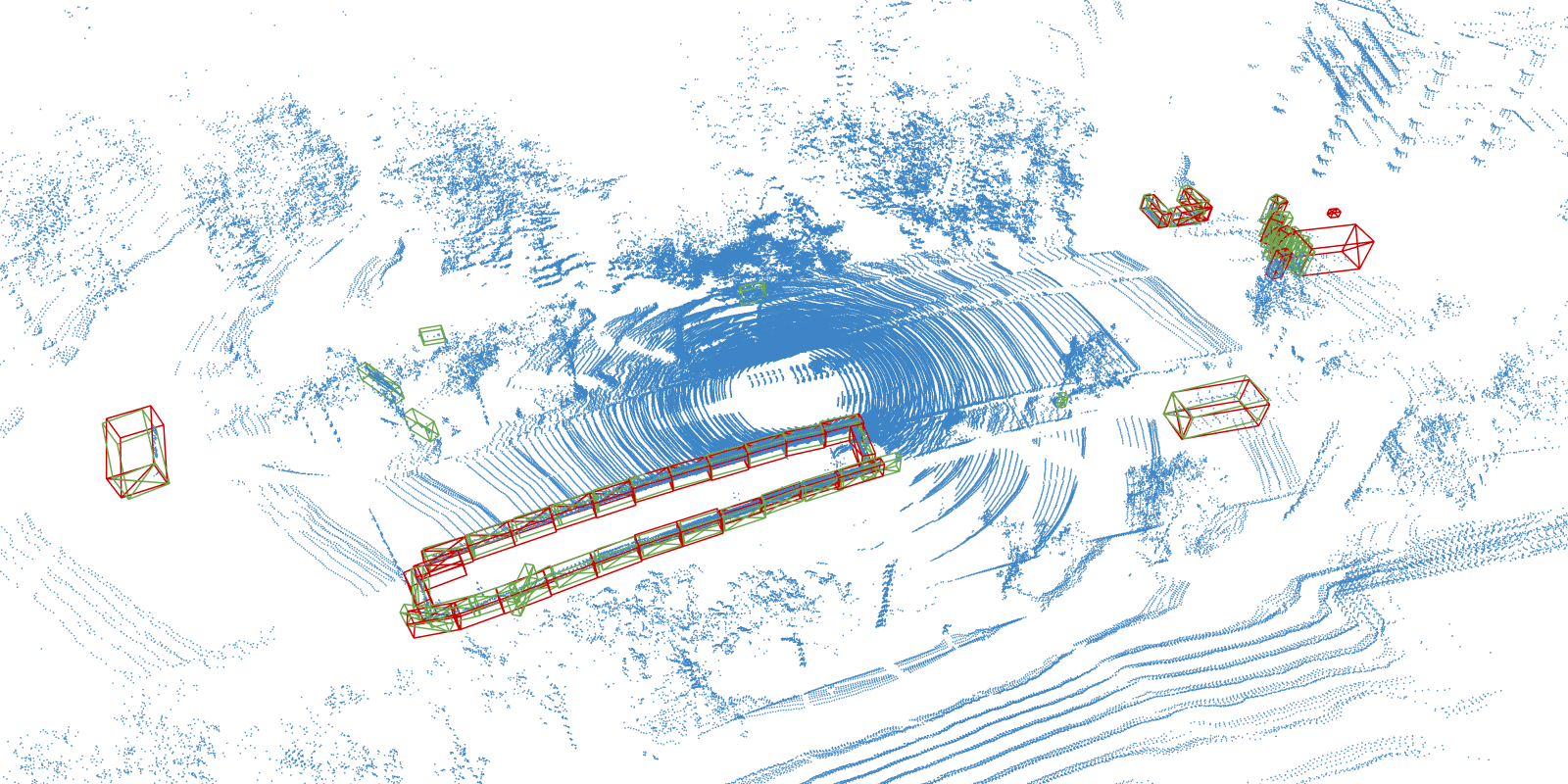} \\
    \includegraphics[width=1.0\textwidth,trim={0cm 1cm 0cm 0cm},clip]{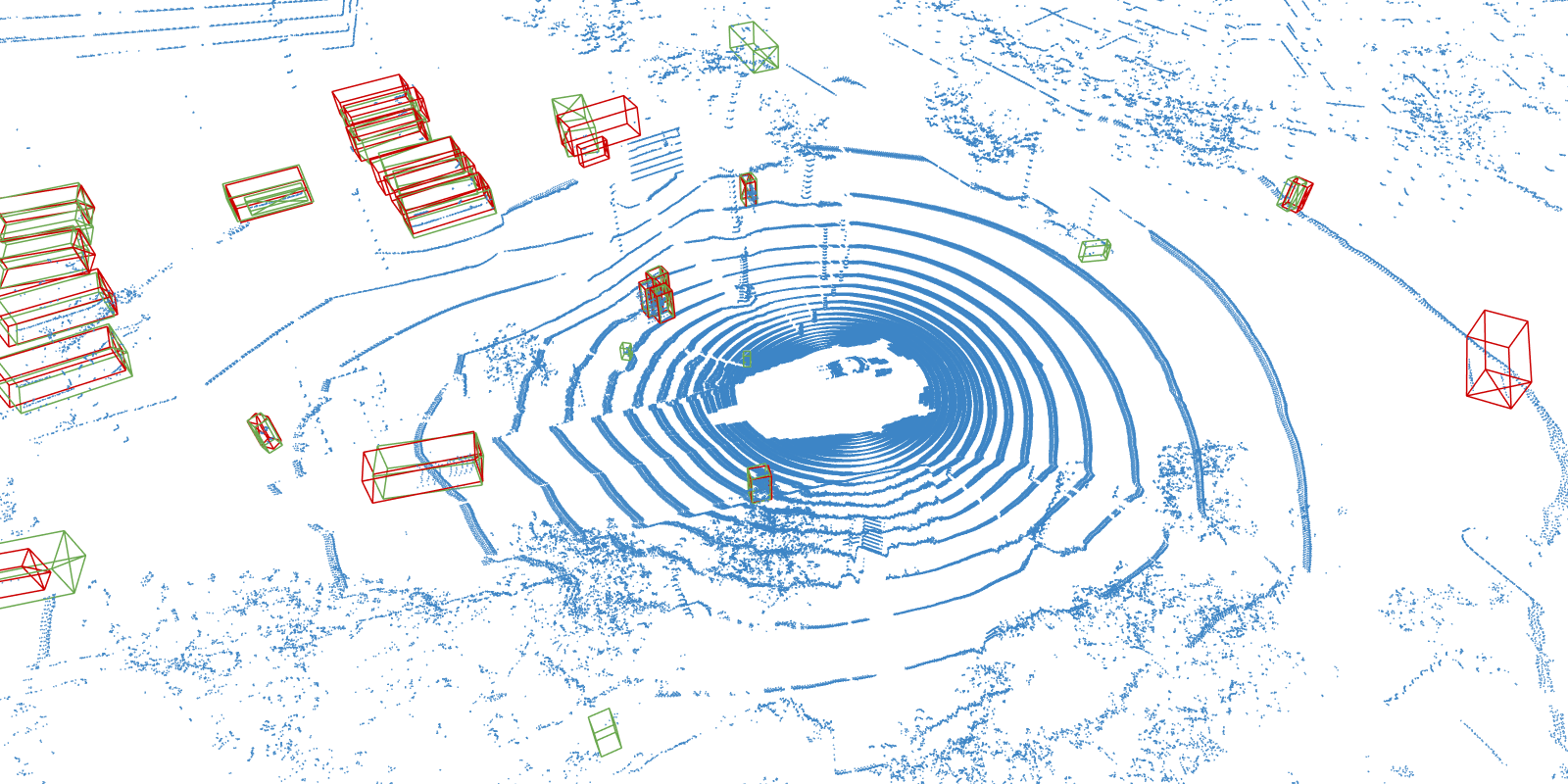} \\
    \includegraphics[width=1.0\textwidth,trim={0cm 0cm 0cm 1.5cm},clip]{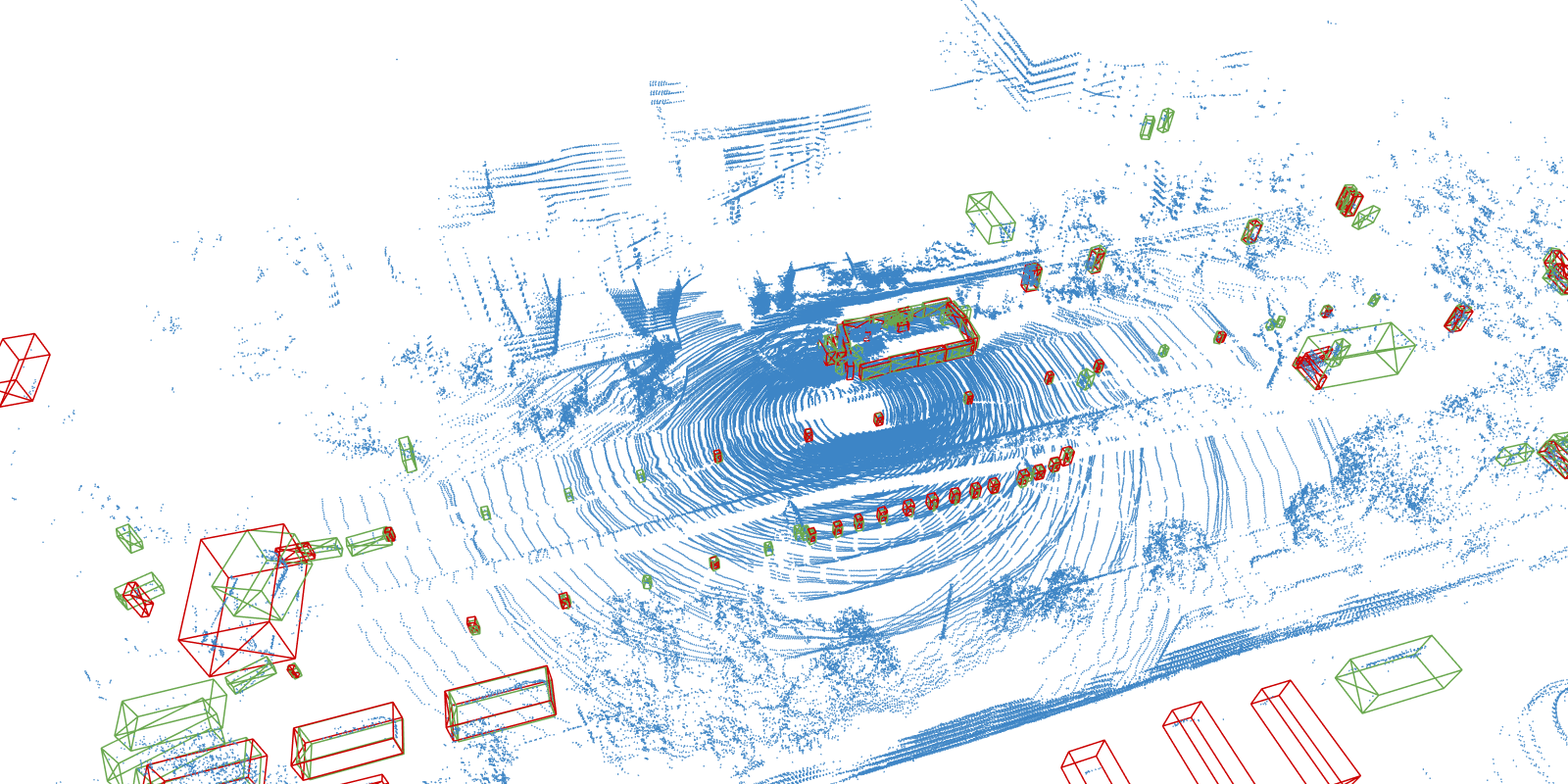}

    \caption{
        Qualitative detection results of CenterPoint~\cite{Yin21CVPR} with GHA (7.5\,cm voxels) on the nuScenes~\cite{Caesar20CVPR} validation set.
        {\color{gs_green} Predictions} and {\color{gs_red} groundtruth} are shown as {\color{gs_green} green} and {\color{gs_red} red} bounding boxes.
        Per convention, each LiDAR scan is an accumulation of five consecutive sweeps~\cite{Caesar20CVPR}.
    }
    \label{fig:nuscenes_qualiative_begin}
\end{figure}

\begin{figure}
    \includegraphics[width=1.0\textwidth,trim={0cm 3cm 0cm 0cm},clip]{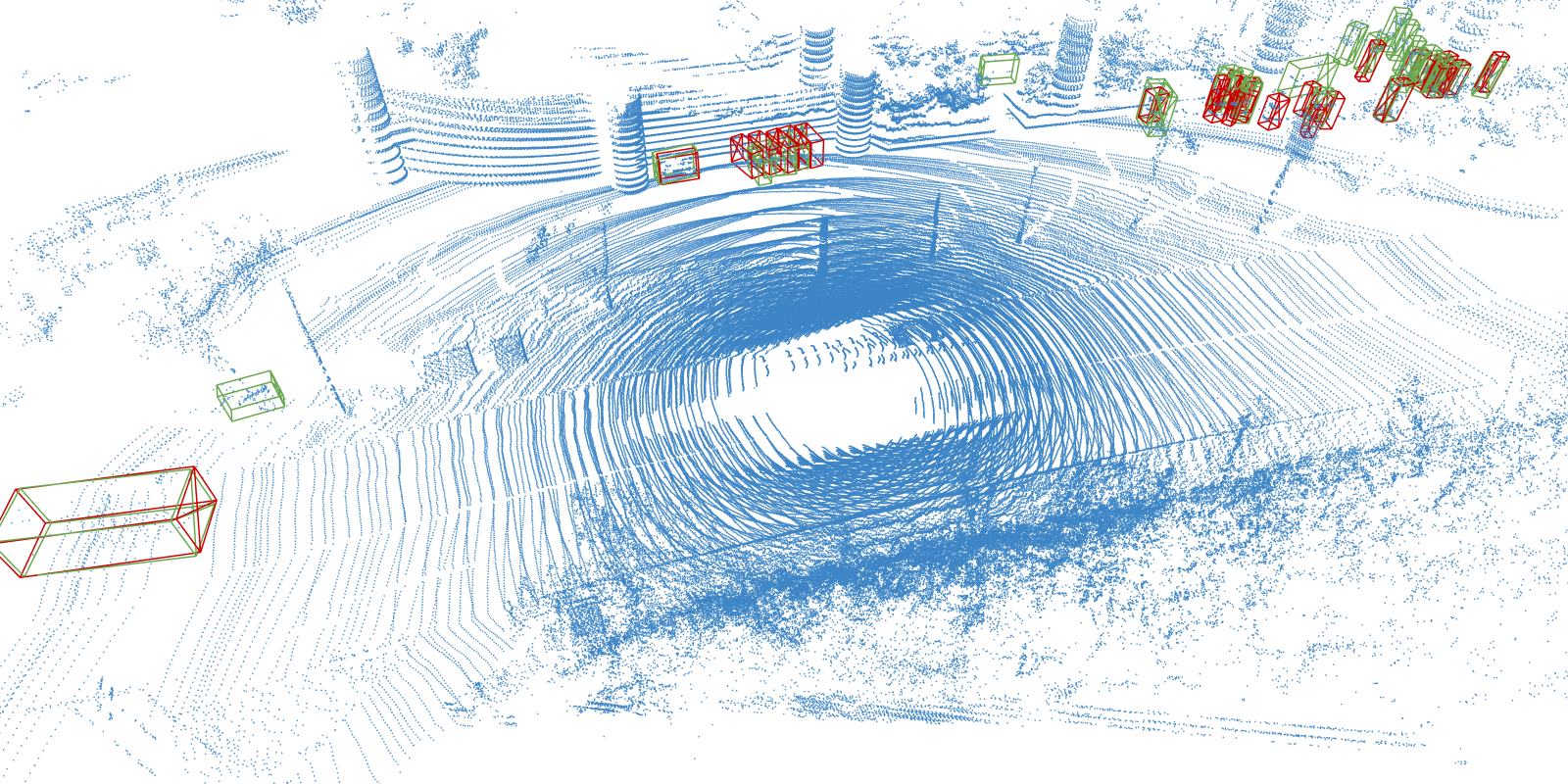} \\
    \includegraphics[width=1.0\textwidth]{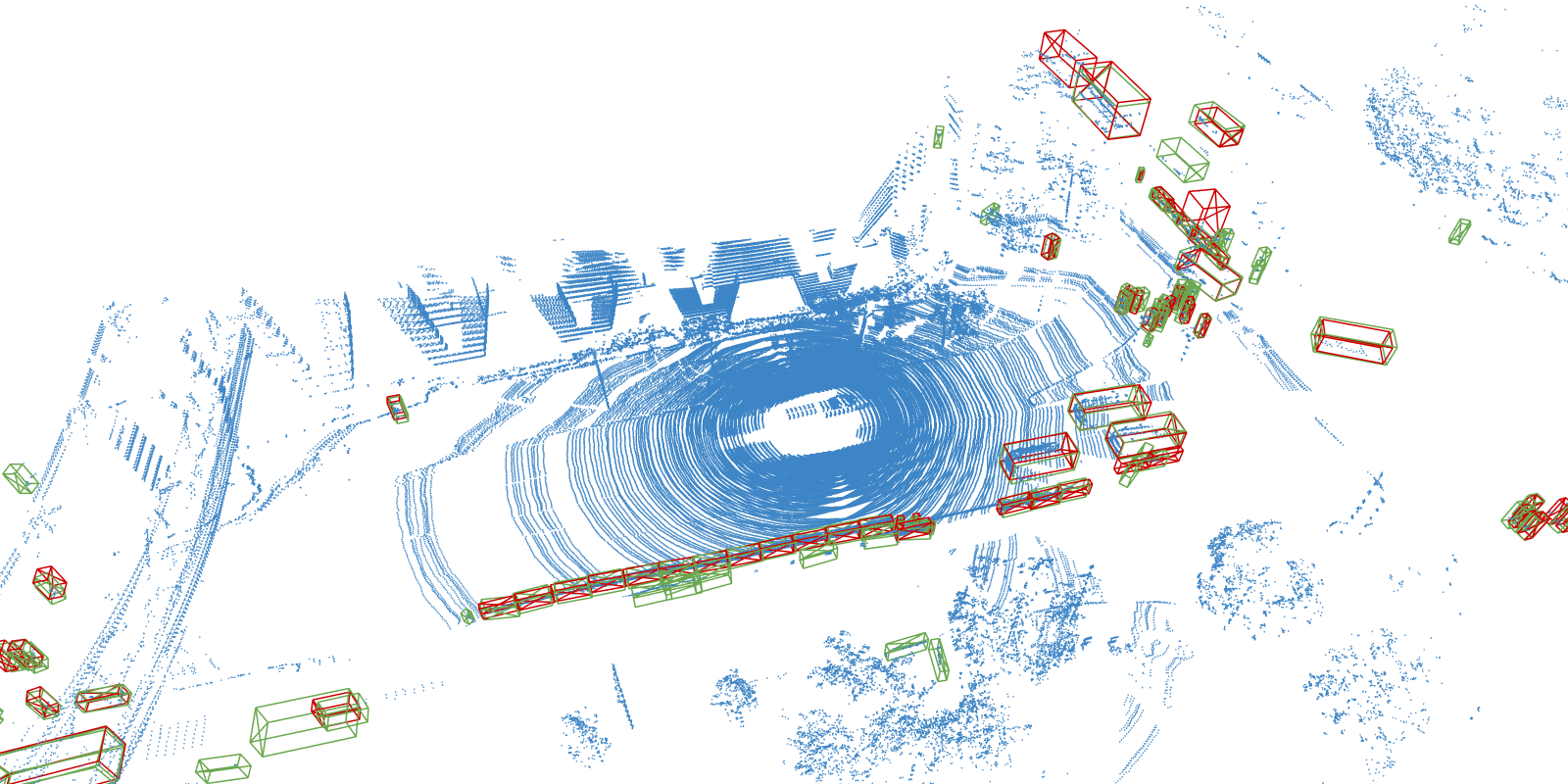} \\
    \includegraphics[width=1.0\textwidth,trim={0cm 0cm 0cm 3cm},clip]{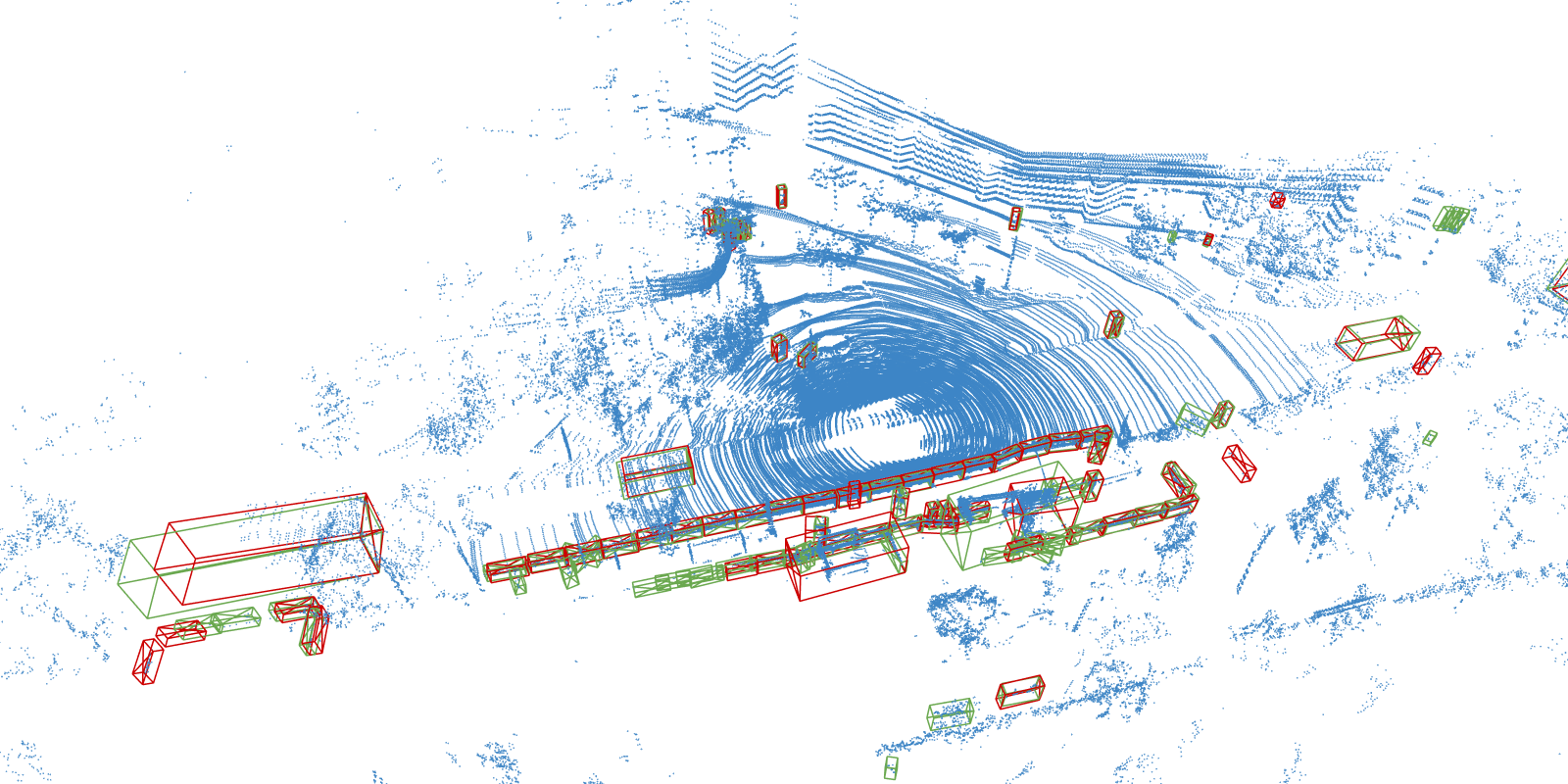}

    \caption{
        Qualitative detection results of CenterPoint~\cite{Yin21CVPR} with GHA (7.5\,cm voxels) on the nuScenes~\cite{Caesar20CVPR} validation set.
        {\color{gs_green} Predictions} and {\color{gs_red} groundtruth} are shown as {\color{gs_green} green} and {\color{gs_red} red} bounding boxes.
        Per convention, each LiDAR scan is an accumulation of five consecutive sweeps~\cite{Caesar20CVPR}.
    }
    \label{fig:nuscenes_qualiative_end}
\end{figure}

In~\reffig{fig:kitti_qual_begin} and \ref{fig:kitti_qual_end} we show qualitative detection results of GHA-augmented SECOND~\cite{Yan18Sensors} on the KITTI validation set.
\reffig{fig:nuscenes_qualiative_begin} and \ref{fig:nuscenes_qualiative_end} shows the results for the GHA-augmented CenterPoint~\cite{Yin21CVPR} on the nuScenes validation set.
Both detectors give good detection results even for small object classes.

\clearpage

{
\bibliographystyle{splncs04}
\bibliography{abbrev_short,mybib}
}

\end{document}